\UseRawInputEncoding

\documentclass[journal]{IEEEtran}

\usepackage{subfigure}
\usepackage{times}
\usepackage{epsfig}
\usepackage{graphicx}
\usepackage{amsmath}
\usepackage{amssymb}
\usepackage{multirow}
\usepackage{color}
\usepackage{cite}
\usepackage{array}
\begin{document}

\title{Asymmetric Hash Code Learning for Remote Sensing Image Retrieval}
\author{Weiwei Song, Zhi Gao, Renwei Dian, Pedram Ghamisi,~\IEEEmembership{Senior Member,~IEEE,} Yongjun Zhang, and
J\'on Atli Benediktsson,~\IEEEmembership{Fellow,~IEEE}
\thanks{ This work was partially supported by the Major Key Project of PCL (Peng Cheng Laboratory) under Grant PCL2021A09, the National Natural Science Foundation of China under Grant 42192583, the Natural Science Foundation of Hubei Province under Grants 2020CFA003 and 2021CFA088, the China Postdoctoral Science Foundation under Grant 2020M682553, and the Fellowship of China National Postdoctoral Program for Innovative Talents under Grant BX20200121.}

\thanks{W. Song is with the Department of Mathematics and Theories, Peng Cheng Laboratory, Shenzhen 518000, China (e-mail: weiweisong415@gmail.com). \par
Z. Gao and Y. Zhang are with the School of Remote Sensing and Information Engineering, Wuhan University, Wuhan 430079, China (e-mail: gaozhinus@whu.edu.cn; zhangyj@whu.edu.cn). \par
R. Dian is with the College of Electrical and Information Engineering,
Hunan University, Changsha 410082, China, also with the Key Laboratory
of Visual Perception and Artificial Intelligence of Hunan Province, Hunan
University, Changsha 410082, China. (e-mail: drw@hnu.edu.cn).  \par
P. Ghamisi is with the Helmholtz-Zentrum Dresden-Rossendorf (HZDR), Helmholtz Institute Freiberg for Resource Technology (HIF), Exploration, D-09599 Freiberg, Germany, and also with the Institute of Advanced Research in Artificial Intelligence (IARAI), Landstra{\ss}er Hauptstra{\ss}e 5, 1030 Vienna, Austria (e-mail: p.ghamisi@gmail.com).  \par
J. A. Benediktsson is with the Faculty of Electrical and Computer Engineering,
University of Iceland, 101 Reykjavk, Iceland (e-mail: benedikt@hi.is).}
}


\maketitle

\begin{abstract}

Remote sensing image retrieval (RSIR), aiming at searching for a set of similar items to a given query image, is a very important task in remote sensing applications. Deep hashing learning as the current mainstream method has achieved satisfactory retrieval performance. On one hand, various deep neural networks are used to extract semantic features of remote sensing images. On the other hand, the hashing techniques are subsequently adopted to map the high-dimensional deep features to the low-dimensional binary codes. This kind of methods attempts to learn one hash function for both the query and database samples in a symmetric way. However, with the number of database samples increasing, it is typically time-consuming to generate the hash codes of large-scale database images. In this paper, we propose a novel deep hashing method, named asymmetric hash code learning (AHCL), for RSIR. The proposed AHCL generates the hash codes of query and database images in an asymmetric way. In more detail, the hash codes of query images are obtained by binarizing the output of the network, while the hash codes of database images are directly learned by solving the designed objective function. In addition, we combine the semantic information of each image and the similarity information of pairs of images as supervised information to train a deep hashing network, which improves the representation ability of deep features and hash codes. The experimental results on three public datasets demonstrate that the proposed method outperforms symmetric methods in terms of retrieval accuracy and efficiency. The source code is available at https://github.com/weiweisong415/Demo\_AHCL\_for\_TGRS2022.

\end{abstract}


%
\IEEEpeerreviewmaketitle

\section{Introduction}

\IEEEPARstart{A}{s} a result of the rapid development of Earth observation technologies, remote sensing images collected by satellites or aerial vehicles have been dramatically enhanced both in volume and resolution. How to effectively manage and analyze these massive amounts of remote sensing images has become an urgent challenge. Remote sensing image retrieval (RSIR), which aims at searching for a set of similar images or scenes to a given query image, has been attracted wide attention in the remote sensing community \cite{PatternNet}.   \par

\begin{figure*}
\begin{center}
\includegraphics[width=150mm]{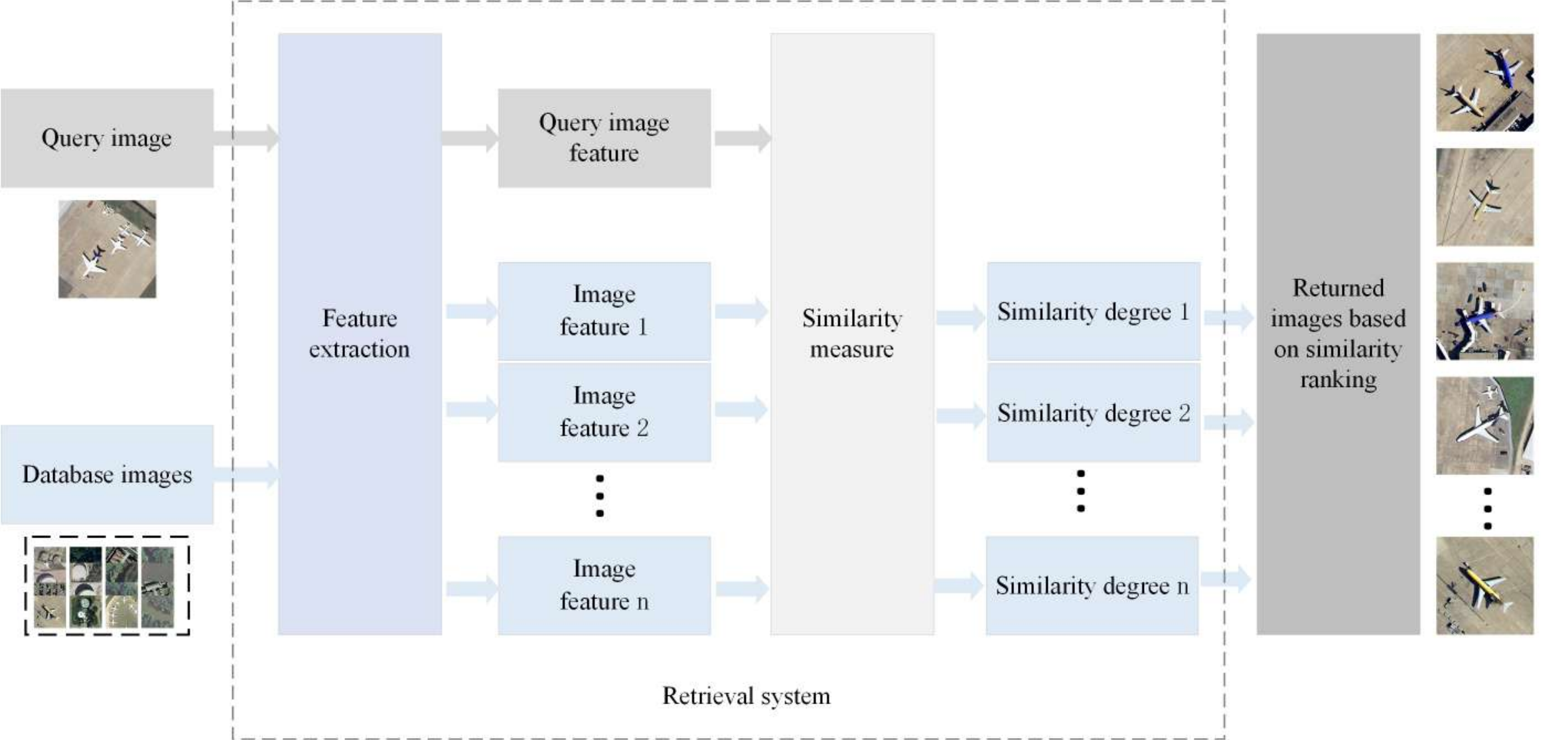}
\end{center}
\caption{Illustration of a typical retrieval framework for RSIR.}
\label{fig:retrieval_framework}
\end{figure*}

In early research of RSIR, most methods exploited annotated tags (e.g., geographical location, acquisition time, or sensor type) to search similar images. Since the used annotated tags can not fully represent image content, this kind of methods usually delivers imprecise retrieval results. By contrast, content-based image retrieval (CBIR) methods employ image features to represent the visual content of remote sensing images, which obtains satisfactory performance. Generally speaking, a CBIR framework includes two main modules: feature extraction and a similarity measure. Fig. \ref{fig:retrieval_framework} demonstrates a typical retrieval framework for RSIR.  \par

For the feature extraction procedure, the query images and database images are all represented by the designed feature descriptors. The extracted features can be divided into hand-crafted features and deep features. The hand-crafted features include low-level features and mid-level features. In the past decades, low-level features were widely used in RSIR, such as texture features \cite{xia-texture}, spectral features \cite{spectral}, and shape features \cite{shape1,shape2}. In addition, various encoding techniques, e.g., bag-of-visual words (BoVW) \cite{BoVW}, Fisher vector (FV) \cite{visual-CNN}, and vector of locally aggregated descriptors (VLAD) \cite{VLAD} were also exploited to encode the low-level features into mid-level features, which delivered satisfactory retrieval results. However, the representation ability of hand-crafted features is limited to accurately describing the semantic information of remote sensing images, which is also called as ``semantic gap''. With the progress of deep learning in the computer visual field, convolutional neural networks (CNNs) have been widely applied in remote sensing applications, including land cover classification \cite{Hong-More-diverse, Hong-Unmixing, HSIC_DL, DFFN, GCN, Hong-Multimodal}, scene recognition \cite{DL_RS, MSCP}, and image fusion \cite{DHIS, Hyper-Multi-fusion, Dian-review, Dian-tensor-train}. In recent years, researchers have also exploited high-level features extracted by CNNs for RSIR and achieved great success \cite{Tong-review, Zhou-HRRSIR, GRN-SNDL-BCE}.  \par

Once the remote sensing image features have been obtained, a similarity measure is subsequently applied to compute the similarity between query and database images. Most of the existing methods adopt Euclidean distance to measure the similarity. However, it is time-consuming for computing the Euclidean distance between two real-valued features, especially for high-dimensional deep features \cite{SPDF}. In order to solve the above problem, hashing techniques have been largely developed for image retrieval \cite{hashing-scalable, Li-PRH}. The main idea behind hashing methods is to learn a set of hash functions that map the high-dimensional image features to low-dimensional hash codes (i.e., binary codes). Different from the complex computation of Euclidean distance, the feature distance between two binary codes (i.e., Hamming distance) can be easily computed via the simple XOR operation.  \par

Recently, deep hashing methods have become the mainstream methods for RSIR. On one hand, deep neural networks are used to extract semantic features for effective content representation. On the other hand, hashing techniques are subsequently adopted to learn binary codes for fast similarity computation. In the past several years, a number of deep hashing methods have been developed for RSIR. For example, Li \emph{et al.} proposed deep hashing neural networks (DHNNs) for large-scale RSIR \cite{DHNN}. Specifically, a pre-trained CNN and a hashing network were exploited to learn high-level semantic features and compact hash codes, respectively. Tang \emph{et al.} embedded hash learning in the generative adversarial framework to ensure the coding balance intuitively \cite{SSDH}. In addition, a cohesion intensive deep hashing model was developed for RSIR, where the cohesiveness of image hash codes within one class was intensified via a weighted loss strategy \cite{CIDH}. In \cite{DH_HPS}, Shan \emph{et al.} combined hash code learning with hard probability sampling in a deep network to improve retrieval performance. In \cite{FAH}, a feature and hash (FAH) learning method, which consists of a deep feature learning model and an adversarial hash learning model, was proposed for RSIR. In \cite{DHCNN}, Song \emph{et al.} proposed a novel deep hashing network simultaneously for remote sensing image retrieval and classification. The above methods attempt to learn one hash function for both query and database samples in a symmetric way. Specifically, the hash codes of query and database images are all obtained by binarizing the output of the network. However, with the number of database samples increasing, the training of this kind of symmetric deep hashing networks becomes typically time-consuming.   \par

To achieve fast image retrieval, Jiang \emph{et al.} \cite{ADSH} proposed an asymmetric deep supervised hashing (ADSH) method to generate the hash codes of query and database images in an asymmetric way. In more detail, the hash codes of query images are obtained via the feedback computation of the deep hashing network, while the hash codes of database images are directly learned by solving the designed objective function. Motivated by \cite{ADSH}, in this paper, we propose a novel asymmetric hashing method named asymmetric hash code learning (AHCL) for RSIR. Different from ADSH that only considered similarity information between image pairs, we elaborately design a better object function which simultaneously combines the semantic information of each image and similarity information between image pairs to train a deep hashing network in an end-to-end way. By fusing multiple kinds of supervised information in object function, our proposed method can extract the more discriminative deep features to represent the complex remote sensing images. The main contributions of this paper can be summarized as follows:
\begin{itemize}
\item{} A novel asymmetric way is developed to generate hash codes of query and database images, respectively. Compared with the existing symmetric deep hashing approaches, the proposed AHCL can significantly improve the efficiency of generating hash codes. To our knowledge, this is the first time to adopt an asymmetric hashing method for RSIR.
\item{} A loss function is elaborately designed to train the propsoed AHCL in an end-to-end way. This loss function combines the semantic loss of each image and similarity loss of pairs of images simultaneously, which improves the representation ability of deep features.
\item{} Comprehensive experiments are performed on three public datasets. The experimental results demonstrate that the proposed AHCL outperforms other competitive deep hashing methods not only in terms of retrieval efficiency but also in terms of retrieval accuracy.
\end{itemize}

The remaining parts of this paper are organized as follows. Section \ref{sec:RW} briefly introduces the related work. Section \ref{sec:methodology} describes in detail different steps of the proposed method. Comprehensive experiments and discussions are exhibited in Section \ref{sec:exp}. Finally, conclusions are presented in Section \ref{sec:conclusions}.\par

\section{Related Work}
\label{sec:RW}
\subsection{Learning to Hash}

Similarity search is a fundamental problem in information retrieval and data mining applications \cite{hash_survey}. With the rapid growth of image data, the search time for similar items is typically expensive or impossible. Approximate nearest neighbor (ANN) search has become a hot research topic in recent years. Among ANN techniques, hashing has become one of the most popular and effective techniques due to its encouraging efficiency in both speed and storage. The goal of hashing is to
learn a set of hash functions that map the image points from the original space into a Hamming space. Through the hashing transformation, each image is represented by a compact binary code and the similarity in the original space is also preserved. Existing learning to hash methods can be roughly divided into two categories: unsupervised hashing and supervised hashing.  \par

For unsupervised hashing methods, the hash functions are learned from unlabeled training data. Spectral hashing (SH) \cite{SH}, iterative quantization (ITQ) \cite{ITQ}, and density sensitive hashing (DSH) \cite{DSH} are the typical unsupervised hashing methods. Due to the low capacity of hash codes, unsupervised hashing methods are usually not robust to noise and image transformations. By contrast, supervised hashing methods try to utilize supervised information to learn hash codes. The supervised information can be given in three different forms: point-wise labels, pairwise labels and ranking labels. The representative supervised hashing methods include supervised hashing with kernels (KSH) \cite{KSH} and sparse embedding and least variance encoding (SELVE) \cite{SELVE}. In addition, a series of deep hashing methods have been developed in the past several years. For example, Xia \emph{et al.} decomposed the hash learning process into a stage of fitting approximate binary codes, followed by a stage of simultaneously fine-tuning the image features and hash functions via a CNN \cite{CNNH}. Li \emph{et al.} adopted pairwise labels information to simultaneously perform feature learning and hash code learning for image retrieval \cite{DPSH}. In addition, Zhang \emph{et al.} utilized pseudo labels to train a deep hashing network in an unsupervised way for scalable image retrieval \cite{Unsupervised-pseudo-labels}. \par

\subsection{Deep Hashing in RSIR}

Traditional RSIR methods exploit hand-crafted features to represent image content. However, the hand-crafted features cannot accurately describe the semantic information of remote sensing images, which delivers suboptimal retrieval results. With the great progress of deep learning in the computer visual field, combining CNNs with hashing techniques has become the mainstream method of RSIR. In the past several years, many related algorithms have been developed. For example, Li \emph{et al.} introduced deep hashing neural networks (DHNNs) for single-source RSIR \cite{DHNN} and cross-source RSIR \cite{source-invariant-DHCNN}, respectively. In \cite{SSDH}, a semi-supervised deep adversarial hashing (SDAH) was proposed for large-scale RSIR tasks. In such a work, a residual auto-encoder (RAE) was used to generate the class variable and hash code. Then, two multi-layer networks were constructed to regularize the above vectors. In \cite{DH_HPS}, Shan \emph{et al.} proposed hard probability sampling hash retrieval method to improve retrieval performance. In \cite{FAH}, Liu \emph{et al.} adopted a deep feature learning model and an adversarial hash learning model to extract dense features of images and map the dense features onto the compact hash codes, respectively. In addition, Song \emph{et al.} designed a unified deep-hashing framework to simultaneously achieve retrieval and classification of remote sensing images \cite{DHCNN}. \par

\section{Proposed Method}
\label{sec:methodology}

\begin{figure*}
\begin{center}
\includegraphics[width=130mm]{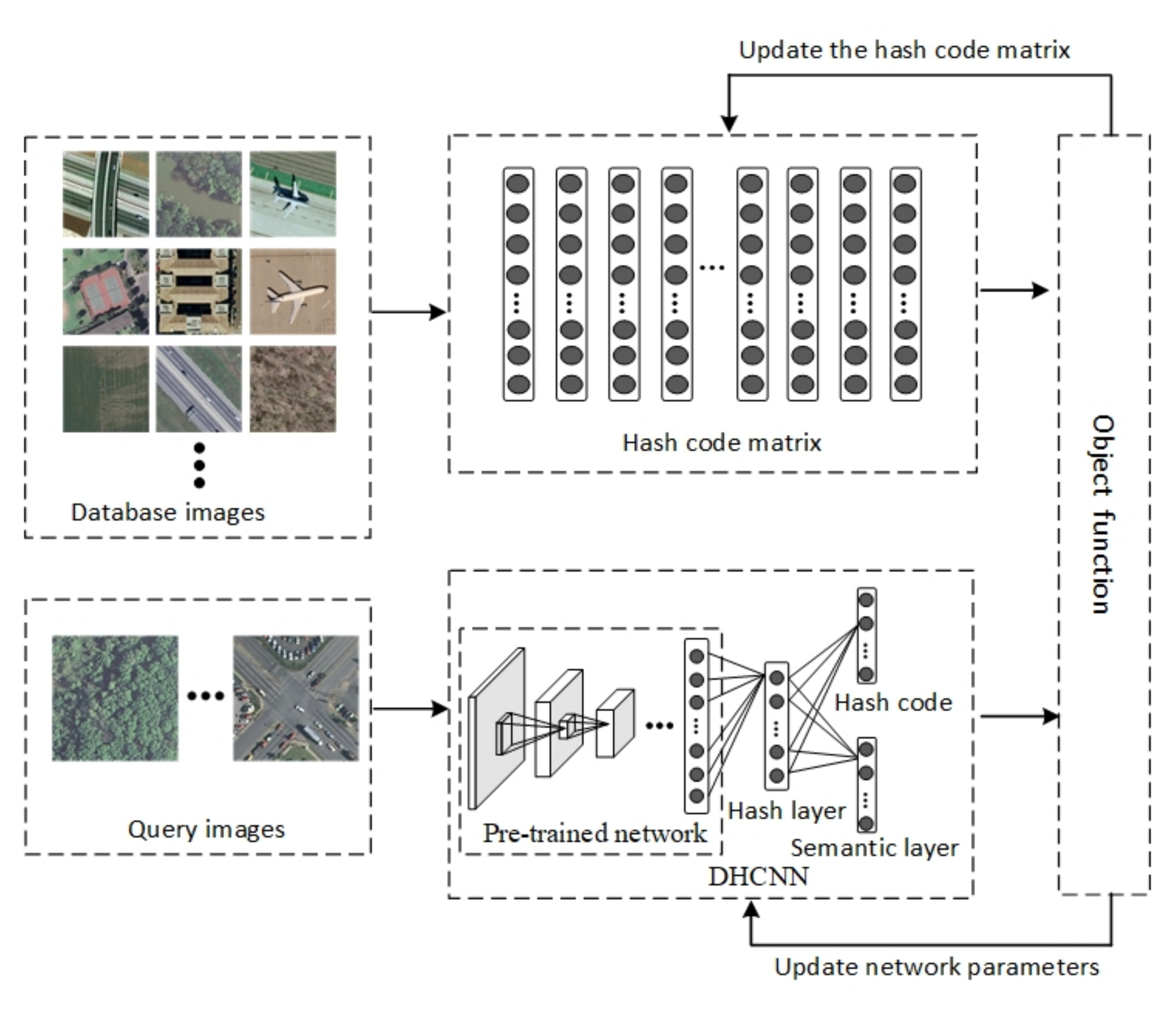}
\end{center}
\caption{The flow chart of the proposed AHCL for RSIR.}
\label{fig:ADH}
\end{figure*}

Recently, a large number of deep supervised hashing methods adopt a symmetric way to learn one hash function for both query images and database images. In more detail, the hash codes of all images are generated by binarizing the output of the network. The retrieval phase of this kind of symmetric deep hashing methods is typically time-consuming due to the repeated feedback computation. \par
To improve the retrieval efficiency, we propose a novel asymmetric deep hashing method for RSIR. Figure~\ref{fig:ADH} presents the schematic of the proposed approach. In the following part, the corresponding procedures are introduced in detail.  \par

\subsection{Deep Feature and Hash Code Extraction}

In this section, we construct a deep hashing convolutional neural network (DHCNN) to simultaneously extract deep features and hash codes of remote sensing images \cite{DHCNN}. The bottom right of Fig.~\ref{fig:ADH} presents the structure of DHCNN, which mainly consists of three parts. First, a pre-trained deep network is transformed as the backbone network, where the last classification layer is discarded. Note that a number of existing successful deep networks can be used as the backbone network, such as CNN-F \cite{CNN-F}, AlexNet \cite{Alexnet}, VGG \cite{VGG}, and ResNet \cite{DRN}. Then, a new fully connected layer named as the hash layer is built on the backbone network to learn the compact hash codes. Here, we denote $K$ as the size of the hash layer. Finally, another fully connected layer with a softmax activation function, named as the semantic layer, is added after the hash layer to generate the label probability distribution. The size of the semantic layer equals the number of classes in a dataset, which denotes as $C$.  \par

Let $\mathbf{Q}=\left\{\mathbf{x}_{q_i}\right\}_{i=1}^m$ and $\mathbf{D}=\left\{\mathbf{x}_{d_j}\right\}_{j=1}^n$ be the query sample dataset and database sample dataset, respectively, where $m$ and $n$ are the number of samples. The corresponding label datasets are represented as $\mathbf{Y}_q=\left\{\mathbf{y}_{q_i}\right\}_{i=1}^m$ and $\mathbf{Y}_d=\left\{\mathbf{y}_{d_j}\right\}_{j=1}^n$, where $\mathbf{y}_{q_i}$ and $\mathbf{y}_{d_j}$ are the one-hot vectors with the dimension of $C\times{1}$. To exploit the supervised information of the labels, the similarity matrix $\mathbf{S}=\left\{s_{ij}\right\}\in\left\{-1,+1\right\}^{m\times{n}}$ between the query images and database images can be defined such that $s_{ij}=1$ if $\mathbf{x}_{q_i}$ and $\mathbf{x}_{d_j}$ come from the same class and $s_{ij}=0$ otherwise. For a query image $\mathbf{x}_{q_i}\in\mathbf{Q}$, we can extract its deep feature (i.e., the output of the backbone network) denoted as $\mathbf{f}_{q_i}$ via the feedback computation of DHCNN
\begin{equation}
\mathbf{f}_{q_i}=\Phi(\mathbf{x}_{q_i}; \theta), i=1,2,...m
\end{equation}
where $\Phi$ is the network function characterized by the parameter $\theta$ existed in the pre-trained network. This feedback computation actually performs a series of nonlinear and linear transformations, including convolution, pooling, and nonlinear mapping. Then, the high-dimensional deep feature is mapped to low-dimensional hash code with $K$ bits via binarizing the output of the hash layer, which can be represented as:
\begin{equation}
\mathbf{b}_{q_i}=sign(\mathbf{u}_{q_i})
\end{equation}
where $\mathbf{u}_{q_i}=\mathbf{W}_h\mathbf{f}_{q_i}+\mathbf{v}_h$ refers to a hash-like code, $\mathbf{W}_h\in\mathbb{R}^{K\times{4096}}$ and $\mathbf{v}_h\in\mathbb{R}^{K\times{1}}$ denote the weight matrix and bias vector of the hash layer, respectively. Here, $sign(\cdot)$ performs an element-wise operation for a matrix or a vector, i.e., $sign(x)=1$ if $x>0$ and $-1$ otherwise.   \par

\subsection{Loss Function Definition}

Due to the complex imaging condition, there exist large intraclass and low interclass variabilities in remote sensing images. To this end, we adopt the product loss between the query images and database images to learn the similarity-preserving deep features, which can be defined as:
\begin{equation}
\begin{split}
\mathcal{L}_1(\mathbf{B}_q, \mathbf{B}_d) = \sum_{i=1}^m \sum_{j=1}^n (\mathbf{b}_{q_i}^T \mathbf{b}_{d_j}-Ks_{ij} )^2
\label{L1}
\end{split}
\end{equation}
where $\mathbf{B}_q=\left\{\mathbf{b}_{q_i}\right\}_{i=1}^m\in\left\{-1, +1\right\}^{m\times{K}}$ and $\mathbf{B}_d=\left\{\mathbf{b}_{d_j}\right\}_{j=1}^n\in\left\{-1, +1\right\}^{n\times{K}}$ represent the hash code matrixes of query image dataset $\mathbf{Q}$ and database image dataset $\mathbf{D}$, respectively. As mentioned above, the hash codes of query images is obtained by DHCNN. Thus, the above Equation (\ref{L1}) can be rewritten as:
\begin{equation}
\begin{split}
\mathcal{L}_2(\mathbf{B}_q, \mathbf{B}_d) = & \sum_{i=1}^m \sum_{j=1}^n [sign(\mathbf{W}_h\Phi(\mathbf{x}_{q_i}; \theta)+\mathbf{v}_h)^T \mathbf{b}_{d_j}\\
                                            & -Ks_{ij} ]^2 .
\end{split}
\label{L2}
\end{equation}
Since the sign function is not derivable, the gradient cannot be prorogated to the former layers of DHCNN. To this end, we adopt the hyperbolic tangent function to approximate the sign function, i.e.:
\begin{equation}
\begin{split}
\mathcal{L}_3(\mathbf{B}_q, \mathbf{B}_d) = & \sum_{i=1}^m \sum_{j=1}^n [tanh(\mathbf{W}_h\Phi(\mathbf{x}_{q_i}; \theta)+\mathbf{v}_h)^T \mathbf{b}_{d_j}\\
                                            & -Ks_{ij} ]^2 .
\end{split}
\label{L3}
\end{equation}
In most cases, the query dataset is randomly sampled from the database dataset, i.e., $\mathbf{B}_q=\mathbf{B}_d^\Omega~$, where $\mathbf{B}_d^\Omega~$ refers to the dataset indicated by the index set $\Omega$. Here, let $\Gamma=\left\{1,2,...,n\right\}$ and $\Omega=\left\{i_1,i_2,...,i_m\right\}\in\Gamma$ be the database sample index set and query sample index set, respectively. Based on the above definition, the loss function is rewritten as:
\begin{equation}
\begin{split}
\mathcal{L}_4(\mathbf{B}_q, \mathbf{B}_d) = & \sum_{i\in\Omega} \sum_{j\in\Gamma} [tanh(\mathbf{W}_h\Phi(\mathbf{x}_{d_i}; \theta)+\mathbf{v}_h)^T \mathbf{b}_{d_j} \\
                                            & -Ks_{ij} ]^2.
\end{split}
\label{L4}
\end{equation}
Considering that $\Omega\in\Gamma$, there are two representation ways of the hash code for query image $\mathbf{x}_{d_i}$. The one is the element of database hash code matrix, i.e., $\mathbf{b}_{d_i}$; the other is the hash-like code (i.e., the output of the hash layer) $\mathbf{\widetilde{u}}_{d_i}=tanh(\mathbf{u}_{d_i})$, where $\mathbf{u}_{d_i}=\mathbf{W}_h\Phi(\mathbf{x}_{d_i}; \theta)+\mathbf{v}_h$. Based on the above analysis, the designed loss function should consider the approximation error. Thus, we add an extra constraint in Equation (\ref{L4}) to make the above two representation ways be as close as possible, i.e.:
\begin{equation}
\begin{split}
\mathcal{L}_5(\mathbf{B}_q, \mathbf{B}_d) = & \sum_{i\in\Omega} \sum_{j\in\Gamma} (\mathbf{\widetilde{u}}_{d_i}^T \mathbf{b}_{d_j}-Ks_{ij} )^2 \\
  & +\lambda\sum_{i\in\Omega}[\mathbf{b}_{d_i}-\mathbf{\widetilde{u}}_{d_i}]^2
\end{split}
\label{L5}
\end{equation}
where $\lambda$ is a hyper-parameter which is used to constrain the representation error.    \par
Equation (\ref{L5}) exploits the similarity information between image pairs to learn similarity-preserving deep features. By minimizing Equation (\ref{L5}), images from the same classes should be encoded as closely as possible and images from the different classes should be encoded far from each other in the feature space. However, apart from information on the similarity between images, each image has its own rich semantic information. To improve the representation ability of features, the semantic information should be considered in the designed loss function. First, the class probability distribution can be computed via the semantic layer with a softmax activation function, i.e.:
\begin{equation}
\mathbf{t}_{d_i}=softmax(\mathbf{W}_s\mathbf{u}_{d_i}+\mathbf{v}_s), \ i\in\Omega
\end{equation}
where $\mathbf{W}_s\in\mathbb{R}^{C\times{K}}$ and $\mathbf{v}_s\in\mathbb{R}^{C\times{1}}$ refer to the weight matrix and bias vector of the semantic layer, respectively. Finally, the cross-entropy loss function is used to minimize the error between the predicted label and the ground-truth label and further added into Equation (\ref{L5}). The final loss function is rewritten as:
\begin{equation}
\begin{split}
\mathcal{L}_6(\mathbf{B}_q, \mathbf{B}_d) & =  \sum_{i\in\Omega} \sum_{j\in\Gamma} (\mathbf{\widetilde{u}}_{d_i}^T \mathbf{b}_{d_j}-Ks_{ij} )^2  \\
 & +\lambda\sum_{i\in\Omega}(\mathbf{b}_{d_i}-\mathbf{\widetilde{u}}_{d_i})^2 + \gamma(\sum_{i\in\Omega}-\mathbf{y}_{d_i}log\mathbf{t}_{d_i})
\label{L6}
\end{split}
\end{equation}
where $\gamma$ is a hyper-parameter which is used to balance the similarity loss and semantic loss.    \par

In Equation (\ref{L6}), the first term is used to preserve the similarity information between images. The second term constrains the approximation error between two representation ways of hash codes of query images. The third term considers the sematic loss of each image between the predicted label and ground-truth label.  \par
\subsection{Objective Function Solving}

After defining the loss function, the objective function can be written as:
\begin{equation}
\begin{aligned}
\mathcal{J}= &\underset{\Theta, \mathbf{B}_d}{\text{min}}\mathcal{L}_6 =  \text{min}\left\{\sum_{i\in\Omega} \sum_{j\in\Gamma} (\mathbf{\widetilde{u}}_{d_i}^T \mathbf{b}_{d_j}-Ks_{ij})^2 \right.  \\
 & + \left. \lambda\sum_{i\in\Omega}(\mathbf{b}_{d_i}-\mathbf{\widetilde{u}}_{d_i})^2 + \gamma(\sum_{i\in\Omega}-\mathbf{y}_{d_i}log\mathbf{t}_{d_i})  \right\}.
\end{aligned}
\label{J}
\end{equation}
In the above objective function, the variables need to be learned including network parameter $\Theta=\left\{\theta, \mathbf{W}_h, \mathbf{v}_h, \mathbf{W}_s, \mathbf{v}_s~\right\}$ and database hash code matrix $\mathbf{B}_d$. Motivated by \cite{ADSH}, we adopt an alternating optimization algorithm to learn the above two variables. More specifically, we solve one variable with the other one fixed. The specific procedures are as follows.
\subsubsection{Solve $\Theta$ with $\mathbf{B}_d$ fixed}

In order to clearly present the solving process, this section deduces the gradient of the objective function to the parameters of the semantic layer (i.e., $\mathbf{W}_s$ and $\mathbf{v}_s$), the parameters of the hash layer (i.e., $\mathbf{W}_h$ and $\mathbf{v}_h$), and the parameters of the pre-trained network (i.e., $\theta$) in turn. First, the partial derivative of the objective function with respect to the predicted class distribution $\mathbf{t}_{d_i}$ is calculated:
\begin{equation}
\frac{\partial{\mathcal{J}}}{\partial{\mathbf{t}_{d_i}}}=\frac{\partial{\mathcal{L}_6}}
{\partial{\mathbf{t}_{d_i}}}=-\gamma\frac{\mathbf{y}_{d_i}}{\mathbf{t}_{d_i}}.
\end{equation}
Then the gradient of the objective function with respect to the parameters of the semantic layer can be calculated:
\begin{equation}
 \frac{\partial{\mathcal{J}}}{\partial{\mathbf{W}_s}}=\frac{\partial{\mathcal{J}}}{\partial{\mathbf{t}_{d_i}}}
 \frac{\partial{\mathbf{t}_{d_i}}}{\partial{\mathbf{o}_{d_i}}}\frac{\partial{\mathbf{o}_{d_i}}}{\partial{\mathbf{W}_s}}
 =\frac{\partial{\mathcal{J}}}{\partial{\mathbf{t}_{d_i}}}\odot\mathbf{t}_{d_i}\odot(\mathbf{y}_{d_i}-\mathbf{t}_{d_i})\mathbf{u}_{d_i}^T
\end{equation}

\begin{equation}
\frac{\partial{\mathcal{J}}}{\partial{\mathbf{v}_s}}=\frac{\partial{\mathcal{J}}}{\partial{\mathbf{t}_{d_i}}}
 \frac{\partial{\mathbf{t}_{d_i}}}{\partial{\mathbf{o}_{d_i}}}\frac{\partial{\mathbf{o}_{d_i}}}{\partial{\mathbf{v}_s}}
 =\frac{\partial{\mathcal{J}}}{\partial{\mathbf{t}_{d_i}}}\odot\mathbf{t}_{d_i}\odot(\mathbf{y}_{d_i}-\mathbf{t}_{d_i})
\end{equation}
where $\mathbf{o}_{d_i}=\mathbf{W}_s\mathbf{u}_{d_i}+\mathbf{v}_s$, the operator $\odot$ represents an element-by-element multiplication. Furthermore, the partial derivative of the objective function with respect to $\mathbf{u}_{d_i}$ can be represented as:
\begin{equation}
\begin{aligned}
\frac{\partial{\mathcal{J}}}{\partial{\mathbf{u}_{d_i}}}= & \left\{[2\sum_{j\in\Gamma}(\mathbf{\widetilde{u}}_{d_i}^T \mathbf{b}_{d_j}-Ks_{ij})\mathbf{b}_{d_j}]+2\lambda[\mathbf{b}_{d_j}-\mathbf{\widetilde{u}}_{d_i}] \right\}   \\
& \odot(1-\mathbf{\widetilde{u}}_{d_i}^2) + \gamma(-\mathbf{W}_s^T(\mathbf{y}_{d_i}-\mathbf{t}_{d_i})).
\end{aligned}
\label{J_u}
\end{equation}
After obtaining the above partial derivatives, the gradient of the objective function with respect to the parameters of the hash layer is further calculated, i.e.:
\begin{equation}
\frac{\partial{\mathcal{J}}}{\partial{\mathbf{W}_h}}=\frac{\partial{\mathcal{J}}}{\partial{\mathbf{u}_{d_i}}}
 \frac{\partial{\mathbf{u}_{d_i}}}{\partial{\mathbf{W}_h}}
 =\frac{\partial{\mathcal{J}}}{\partial{\mathbf{u}_{d_i}}}\mathbf{f}_{d_i}^T
\end{equation}
\begin{equation}
\frac{\partial{\mathcal{J}}}{\partial{\mathbf{v}_h}}=\frac{\partial{\mathcal{J}}}{\partial{\mathbf{u}_{d_i}}}
 \frac{\partial{\mathbf{u}_{d_i}}}{\partial{\mathbf{v}_h}}
 =\frac{\partial{\mathcal{J}}}{\partial{\mathbf{u}_{d_i}}}.
\end{equation}
Finally, the gradient of the objective function with respect to the parameters of the pre-trained network is also calculated:
\begin{equation}
\begin{split}
\frac{\partial{\mathcal{J}}}{\partial{\theta}} & =
\frac{\partial{\mathcal{J}}}{\partial{\mathbf{u}_{d_i}}}
\frac{\partial{\mathbf{u}_{d_i}}}{\partial{\Phi(\mathbf{x}_{d_i}; \theta)}}
\frac{\partial{\Phi(\mathbf{x}_{d_i};\theta)}}{\partial{\theta}} \\
& =\mathbf{W}_h^T\frac{\partial{\mathcal{J}}}{\partial{\mathbf{u}_{d_i}}}
\frac{\partial{\Phi(\mathbf{x}_{d_i};\theta)}}{\partial{\theta}}.
\end{split}
\end{equation}
When the gradients of all parameters are obtained, the standard gradient descent algorithm (SGD) is used to update all parameters, i.e.:
\begin{equation}
\xi = \xi - \mu\frac{\partial{\mathcal{J}}}{\partial{\xi}}, \xi=\mathbf{W}_s, \mathbf{W}_h, \mathbf{v}_s, \mathbf{v}_h, \theta
\end{equation}
where $\mu$ refers to the learning rate.   \par

\subsubsection{Solve $\mathbf{B}_d$ with $\Theta$ fixed}

When $\Theta$ is fixed, Equation (\ref{J}) can be rewritten in a matrix form, i.e.:
\begin{equation}
\begin{split}
\mathcal{J}= \underset{\mathbf{B}_d}{\text{min}}\mathcal{L}_6 =  & \underset{\mathbf{B}_d}{\text{min}}\{ \|\mathbf{\widetilde{U}}_d \mathbf{B}_d^T-K\mathbf{S}\|_F^2 \\
 & +\lambda\|\mathbf{B}_d^\Omega-\mathbf{\widetilde{U}}_d\|_F^2+
\gamma(-\mathbf{Y}_d log\mathbf{T}_d) \} \\
= &\underset{\mathbf{B}_d}{\text{min}} \{ \|\mathbf{\widetilde{U}}_d \mathbf{B}_d^T\|_F^2-2Ktr(\mathbf{B}_d^T\mathbf{S}^T\mathbf{\widetilde{U}}_d)\\
& - 2\lambda tr(\mathbf{B}_d^\Omega\widetilde{U}_d^T)+const\}.
\label{J_matrix}
\end{split}
\end{equation}
where $\mathbf{\widetilde{U}}_d$, $\mathbf{B}_d$, $\mathbf{S}$, $\mathbf{Y}_d$, and $\mathbf{T}_d$ are the matrix form of the corresponding variables. $\mathbf{B}_d^\Omega$ denotes the hash code matrix for the samples in the database indexed by $\Omega$. ``const'' represents a constant independent of $\mathbf{B}_d$. After that, we define a new variable $\mathbf{\overline{U}}_d=\{\mathbf{\overline{u}}_{d_j}\}_{j=1}^n$, i.e.:
\begin{equation}
\mathbf{\overline{u}}_{d_j}=
\left\{
\begin{array}{lr}
\mathbf{\widetilde{u}}_{d_j},   & if j\in\Omega \\
   0,                           & otherwise.
\end{array}
\right.
\label{u_new}
\end{equation}
Thus, Equation (\ref{J_matrix}) can be rewritten as:
\begin{equation}
\begin{aligned}
\mathcal{J}= &\underset{\mathbf{B}_d}{\text{min}} \{ \|\mathbf{B}_d\widetilde{\mathbf{U}}_d^T\|_F^2-2tr(\mathbf{B}_d[K\mathbf{\widetilde{U}}_d^T\mathbf{S}+
\lambda\mathbf{\overline{U}}_d]^T)+const \}   \\
= &\underset{\mathbf{B}_d}{\text{min}} \{ \|\mathbf{B}_d\widetilde{\mathbf{U}}_d^T\|_F^2+tr(\mathbf{B}_d\mathbf{Q}_d^T) +const \}
\label{J_matrix_new}
\end{aligned}
\end{equation}
where $\mathbf{Q}_d=-2K\mathbf{S}^T\mathbf{B}_d\mathbf{\widetilde{U}}_d-2\lambda\mathbf{\overline{U}}_d$. \par
Based on the above definition, we learn the whole $\mathbf{B}_d$ by updating one column of $\mathbf{B}_d$ successively and fixing the other columns. Assume that $\mathbf{B}_{d_{*k}}$, $\mathbf{\widetilde{U}}_{d_{*k}}$, and $\mathbf{Q}_{d_{*k}}$ are the $k$th columns of $\mathbf{B}_d$, $\mathbf{\widetilde{U}}_d$, and $\mathbf{Q}_d$, respectively, and that $\mathbf{\hat{B}}_{d_k}$, $\mathbf{\hat{U}}_{d_k}$, and $\mathbf{\hat{Q}}_{d_k}$ are the matrices of $\mathbf{B}_d$, $\mathbf{\widetilde{U}}_d$, and $\mathbf{Q}_d$ excluding the $k$th columns, respectively. The objective function can be further rewritten as:
\begin{equation}
\begin{aligned}
\mathcal{J}= & \underset{\mathbf{B}_{d_{*k}}}{\text{min}} \{ \|\mathbf{B}_d\widetilde{U}_d^T\|_F^2+tr(\mathbf{B}_d\mathbf{Q}_d^T) +const \}   \\
= & \underset{\mathbf{B}_{d_{*k}}}{\text{min}} \{tr(\mathbf{B}_{d_{*k}}[2\mathbf{\widetilde{U}}_{d_{*k}}^T\mathbf{\hat{U}}_{d_k}
\mathbf{\hat{B}}_{d_k}^T+\mathbf{Q}_{d_{*k}}^T])+const \}.
\label{J_matrix_new_1}
\end{aligned}
\end{equation}
By solving Equation (\ref{J_matrix_new_1}), we can gradually update $\mathbf{B}_{d_{*k}}$, i.e.:
\begin{equation}
\mathbf{B}_{d_{*k}}=-sign(2\mathbf{\hat{B}}_{d_k}\mathbf{\hat{U}}_{d_k}^T\mathbf{\widetilde{U}}
_{d_{*k}}+\mathbf{Q}_{d_{*k}}).
\end{equation}
\par

After training, for a unseen query image $\mathbf{f}_{q_i}$, its hash code is obtained by using the following equation:
\begin{equation}
\mathbf{b}_{q_i}=sign(\mathbf{u}_{q_i})=sign(\mathbf{W}_h\Phi(\mathbf{x}_{q_i}; \theta)+\mathbf{v}_h).
\end{equation}

\section{Experiments}
\label{sec:exp}
To verify the effectiveness of the proposed method for RSIR, we compare the proposed AHCL against some state-of-the-art methods on three public remote sensing image datasets. In the following part, Section \ref{sec:datasets} introduces the used datasets. Section \ref{sec:settsings} describes the experimental settings. Section \ref{sec:results} reports the experimental results. Section \ref{sec:analysis} discusses the effects of two important parameters of the proposed method on retrieval results. \par

\subsection{Datasets}
\label{sec:datasets}

\begin{figure*}[htbp]
\centering
\includegraphics[width=130mm]{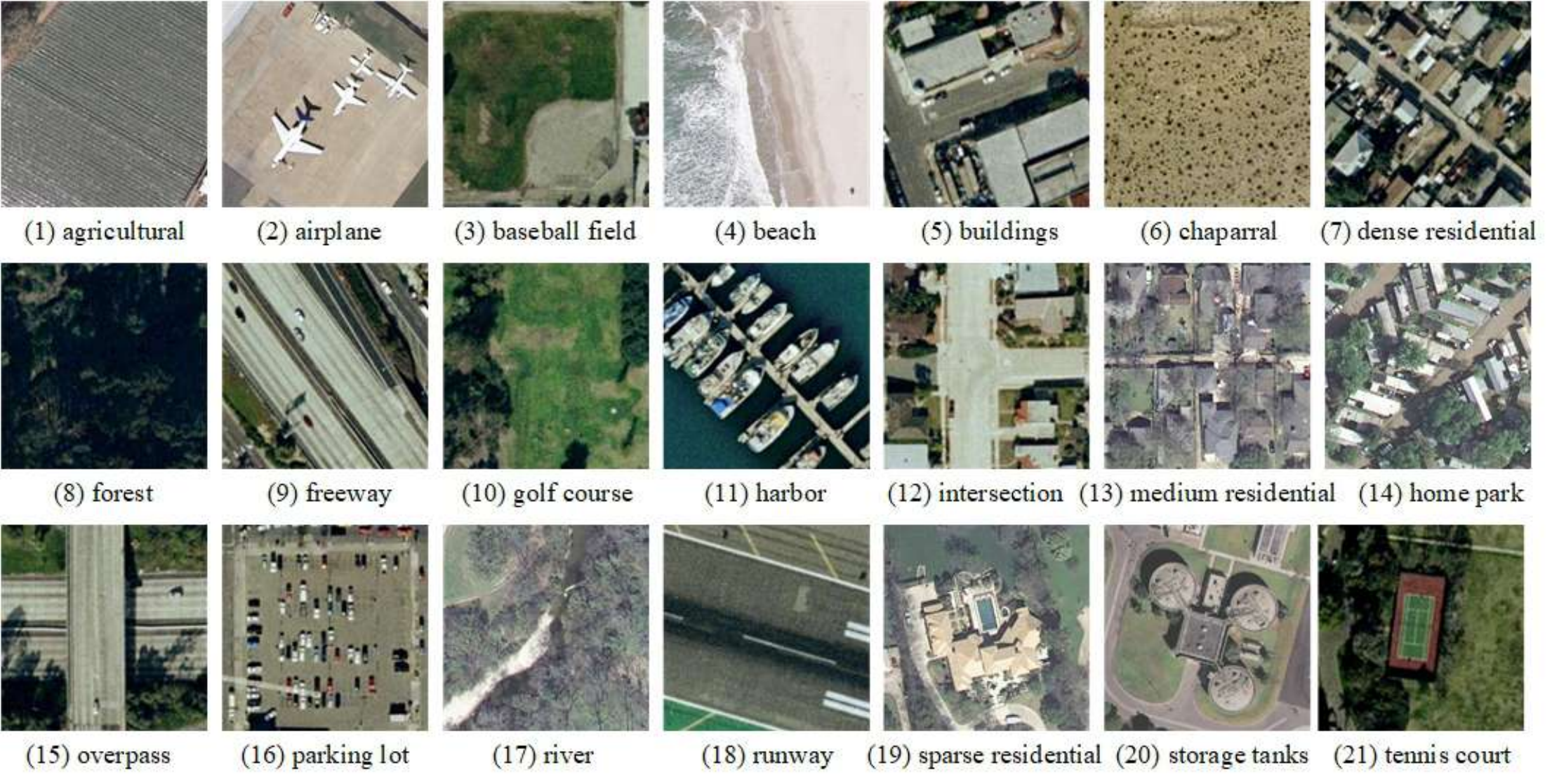}
\caption{Examples of different scenes in the~UCMD.}
\label{fig:UCMD_samples}
\end{figure*}

\begin{figure*}[htbp]
\centering
\includegraphics[width=130mm]{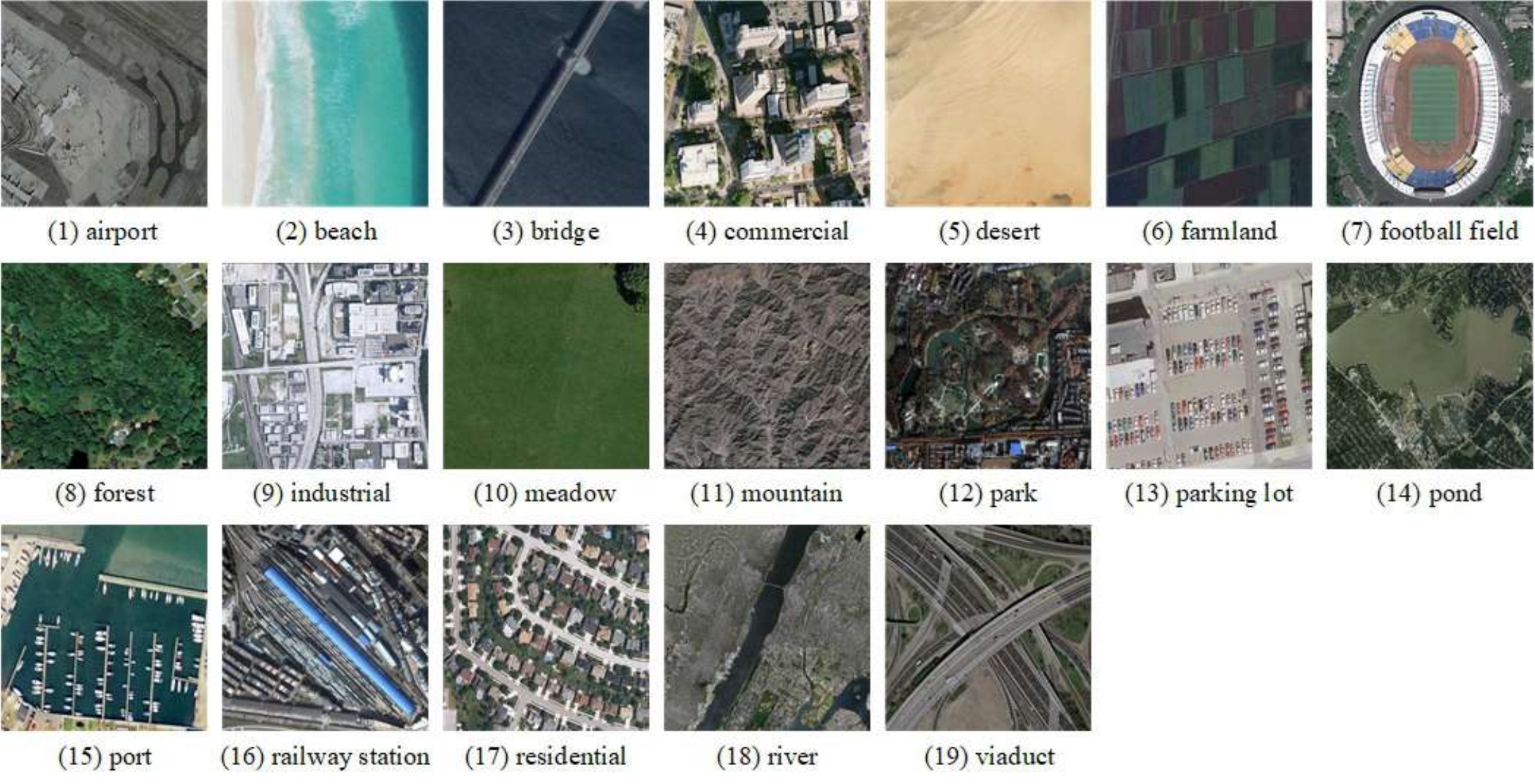}
\caption{Examples of different scenes in the~WHU-RS.}
\label{fig:WHURS_samples}
\end{figure*}

\begin{figure*}[htbp]
\centering
\includegraphics[width=130mm]{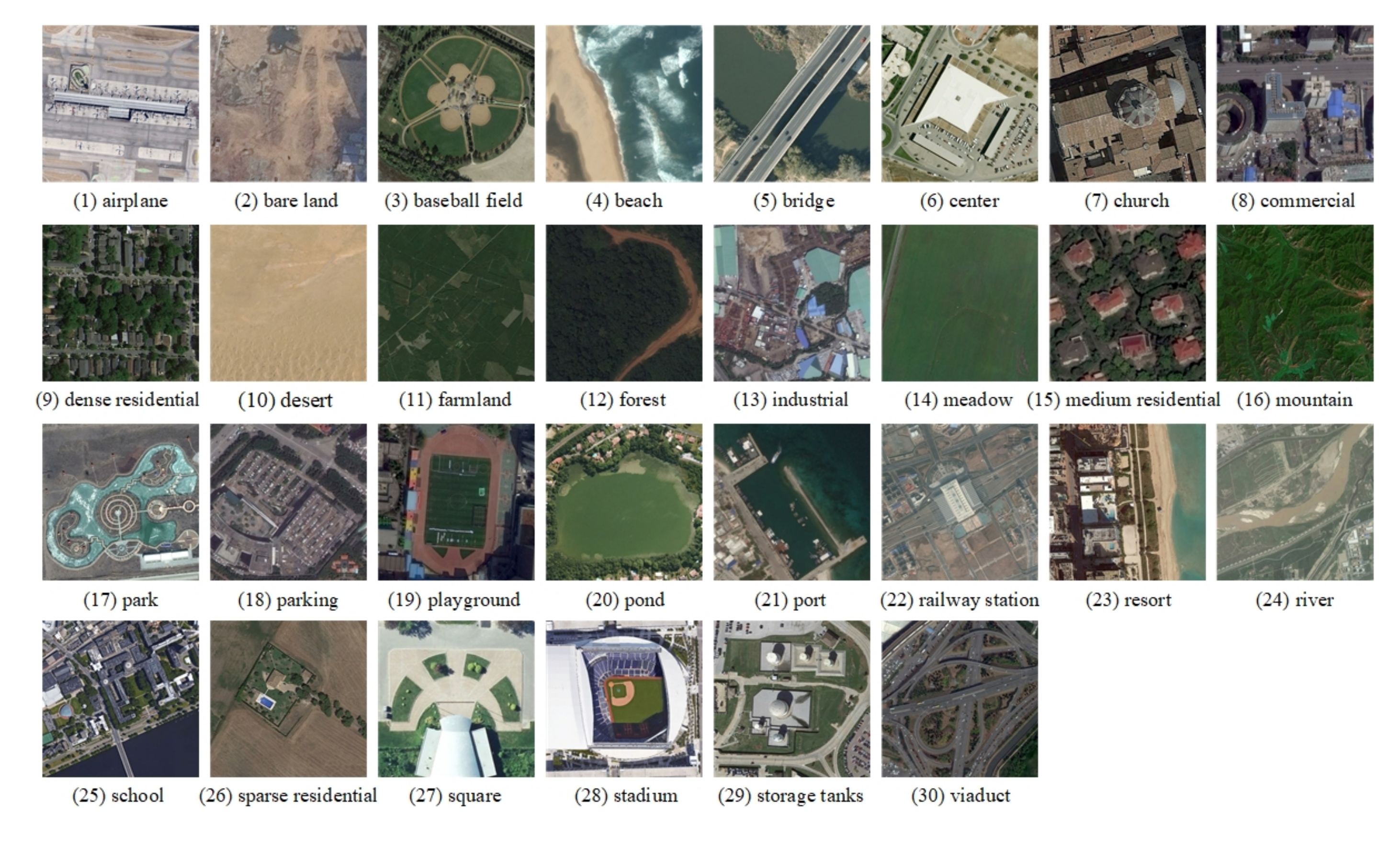}
\caption{Examples of different scenes in the~AID.}
\label{fig:AID_samples}
\end{figure*}
We select three remote sensing image datasets to conduct RSIR experiments. The detailed descriptions of these datasets are as follows:
\begin{itemize}
\item{} The first dataset is the University of California, Merced dataset (UCMD) \cite{UCM21} which was extracted from the United States Geological Survey (USGS). It contains 21 land cover categories and each category includes 100 images. The size of the images is $256\times{256}$, and the spatial resolution of each pixel is 0.3 m. Some examples of different scenes are presented in Fig. \ref{fig:UCMD_samples}. \par

\item{} The second dataset is the WHU-RS dataset \cite{WHU-RS19} which was collected from Google Earth. The images are divided into 19 classes, each class has approximately 50 images with $600\times{600}$ pixels. Some examples of different scenes are presented in Fig. \ref{fig:WHURS_samples}.  \par

\item{} The third dataset is the aerial image dataset (AID) \cite{AID} which was collected with the goal of advancing the state-of-the-art for the scene classification of remote sensing images. The dataset has a number of 10000 images within 30 classes. Each class consists of 220 to 420 images of size of $600\times{600}$ pixels. Some examples of different scenes are presented in Fig. \ref{fig:AID_samples}. \par
\end{itemize}

For UCMD, WHU-RS, and AID datasets, we randomly select 80\%, 50\%, and 50\% of the labeled samples per class as training samples, respectively, the rest of the samples are regarded as the test set.  \par

\subsection{Experimental Settings}
\label{sec:settsings}

\begin{figure*}[htbp]
\begin{center}
\includegraphics[width=180mm]{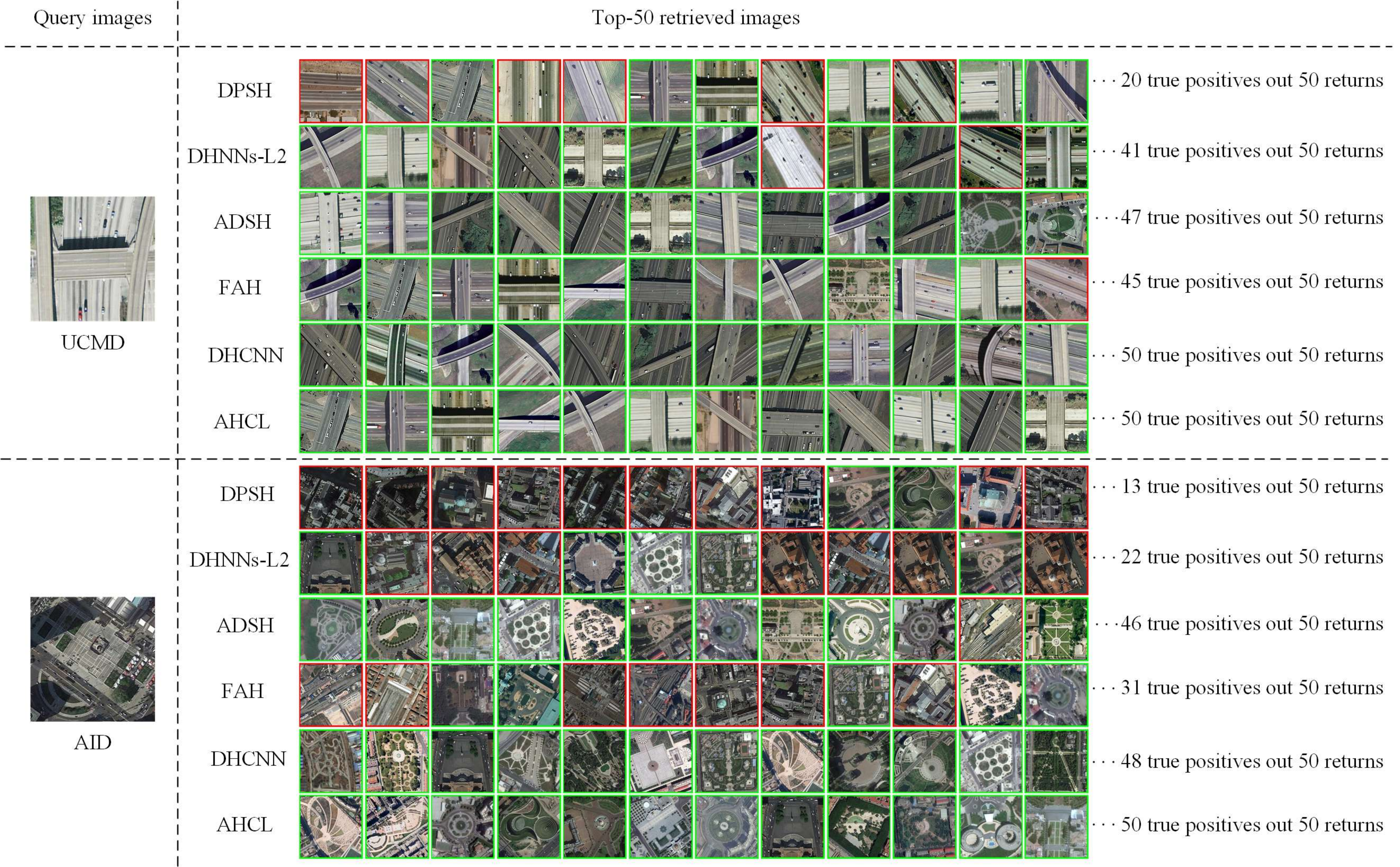}
\caption{The query examples with the top-50 retrieved images on the UCMD and AID datasets, where the green rectangle marks the true positives and the red rectangle marks the false positives.}
\label{fig:retrieval_image}
\end{center}
\end{figure*}

To extract remote sensing image deep features, we adopt VGG11 \cite{VGG} as the pre-trained network (i.e., backbone network), which includes eight convolutional layers and three fully connected layers. In experiments, the last classification layer of VGG11 is discarded. We systematically compare our method with some traditional hashing methods and deep hashing methods. The traditional methods include SELVE \cite{SELVE} and KSH \cite{KSH}. The deep hashing methods include deep pairwise-supervised hashing (DPSH) \cite{DPSH}, and DHNNs with the L2 regularization (DHNNs-L2) \cite{DHNN}, asymmetric deep supervised hashing (ADSH) \cite{ADSH}, FAH \cite{FAH}, and DHCNN \cite{DHCNN}. In addition, another approach named graph relation network with scalable neighbor discriminative loss with binary cross entropy (GRN-SNDL-BCE) \cite{GRN-SNDL-BCE} is also regarded as baseline. For traditional methods, each remote sensing image is represented by 4096-dimensional CNN features extracted from the penultimate layer of VGG11. For deep hashing methods, all images are first resized to be of $224\times{224}$ pixels and then directly fed into the deep networks. The parameters of the compared methods are set to default values according to the original papers. For our proposed AHCL method, the parameters $\lambda$ and $\gamma$ are set to 200 and 20, respectively. All experiments are performed on a computer equipped with an Intel Core i7-9700 with 3.0 GHz, 64G memory, and an NVIDIA GeForce RTX 2060 SUPER GPU.  \par

To evaluate the performance of the retrieval methods, we adopt four metrics, i.e., Mean Average Precision (MAP), Precision@k, Recall@k, and Precision-Recall. The descriptions of these metrics are as follows:
\begin{itemize}
\item{} MAP: In the query phase, we firstly rank all database samples in ascending order by computing the Hamming distance between the query sample and the database samples. Once obtaining the ranked list, we can get the average precision (AP) for each query image. Finally, the MAP can be computed via averaging the AP of all query images, which is defined as:
\begin{equation}
\text{MAP}=\frac{1}{|Q|}\sum_{i=1}^{|Q|}\frac{1}{n_i}\sum_{j=1}^{n_i}P(i,j)
\end{equation}
where $|Q|$ is the volume of the query image set, $n_i$ is the number of images relevant to $i$th query image in the searching database, and $P(i,j)$ is the precision of the top $j$th retrieved
image of $i$th query image. \par

\item{} Precision@k: This metric measures the precision value of the top $k$ retrieved images, which is defined as:
\begin{equation}
\text{Precision@k} = \frac{n}{k}
\end{equation}
where $k$ and $n$ are the numbers of all images and similar images to the query image in the top $k$ list, respectively.  \par

\item{} Recall@k: Recall@k computes the recall rate between the number of similar images to the query image in the top $k$ retrieved image and all similar images in the database, which is defined as:
\begin{equation}
\text{Recall@k} = \frac{n}{r}
\end{equation}
where $r$ and $n$ are the number of similar images in the database and the top $k$ retrieved images, respectively.   \par

\item{} Precision-Recall: The Precision-Recall metric is another popular evaluation protocol in image retrieval, which plots the precision and recall rates at different searching Hamming radius. The first point of the Precision-Recall curve represents the precision and recall rate at the Hamming radius equals 0; the next point means the precision and recall rate at the Hamming radius equals 1, and so on. \par
\end{itemize}

\begin{table*}[htbp]
\small
\centering
\caption{Image Retrieval Results in Terms of MAP With 16, 32, and 64 Hash Bits on the Three Datasets. Note that GRN-SNDL-BCE\cite{GRN-SNDL-BCE_2} Is Not Hashing-based Method, We Set the Length of Feature to 64 for Comparison.}
\begin{tabular}{cccccccccc}
\hline
\multirow{2}{*}{Methods} & \multicolumn{3}{c}{UCMD} & \multicolumn{3}{c}{WHU-RS}  & \multicolumn{3}{c}{AID}   \\
\cline{2-10}
   & 16 bits   & 32 bits & 64 bits & 16 bits & 32 bits & 64 bits & 16 bits & 32 bits & 64 bits  \\
\hline
\textbf{AHCL} (Our method)         & \textbf{0.9709} & \textbf{0.9762} & \textbf{0.9854}
                       & \textbf{0.9661} & \textbf{0.9811} & \textbf{0.9843}
                       & \textbf{0.8990} & \textbf{0.9537} & \textbf{0.9559} \\
DHCNN\cite{DHCNN}      & 0.9682 & 0.9718 & 0.9822 & 0.9412 & 0.9694 & 0.9743
                       & 0.8935 & 0.9457 & 0.9502 \\
GRN-SNDL-BCE\cite{GRN-SNDL-BCE}      & - & - & 0.9833 & - & - & 0.9808
                       & - & - & 0.9506 \\
FAH\cite{FAH}          & 0.9010 & 0.9561 & 0.9653 & 0.7776 & 0.9508 & 0.9649
                       & 0.8494 & 0.9248 & 0.9281 \\

ADSH\cite{ADSH}        & 0.9651 & 0.9689 & 0.9810 & 0.9334 & 0.9494 & 0.9739
                       & 0.8898 & 0.9472 & 0.9493 \\
DHNNs-L2\cite{DHNN}    & 0.9232 & 0.9569 & 0.9649 & 0.8923 & 0.9243 & 0.9502
                       & 0.8239 & 0.8632 & 0.9221 \\
DPSH\cite{DPSH}        & 0.8382 & 0.9135 & 0.9225 & 0.7245 & 0.7941 & 0.8532
                       & 0.7532 & 0.8037 & 0.8822 \\
KSH-CNN\cite{KSH}      & 0.7755 & 0.8475 & 0.8792 & 0.6953 & 0.7532 & 0.8073
                       & 0.5043 & 0.6053 & 0.6531 \\
SELVE-CNN\cite{SELVE}  & 0.3863 & 0.4254 & 0.4308 & 0.4238 & 0.4929 & 0.5032
                       & 0.3508 & 0.3907 & 0.3840 \\

\hline
\end{tabular}
\label{tab:Results-MAP}
\end{table*}

\subsection{Retrieval Results}
\label{sec:results}

\begin{figure*}[htbp]
\begin{center}
\subfigure[]{\includegraphics[width=50mm]{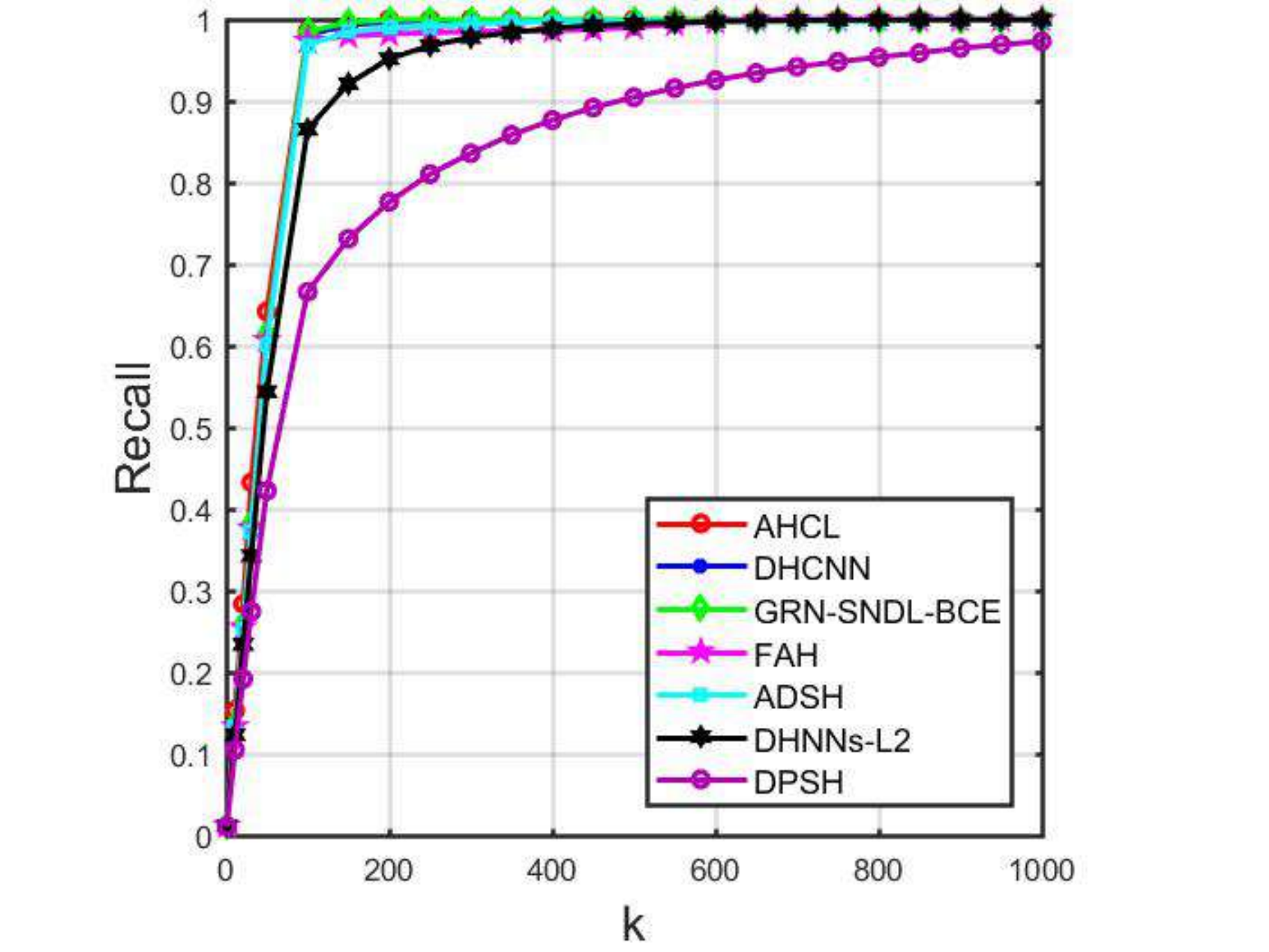}}
\subfigure[]{\includegraphics[width=50mm]{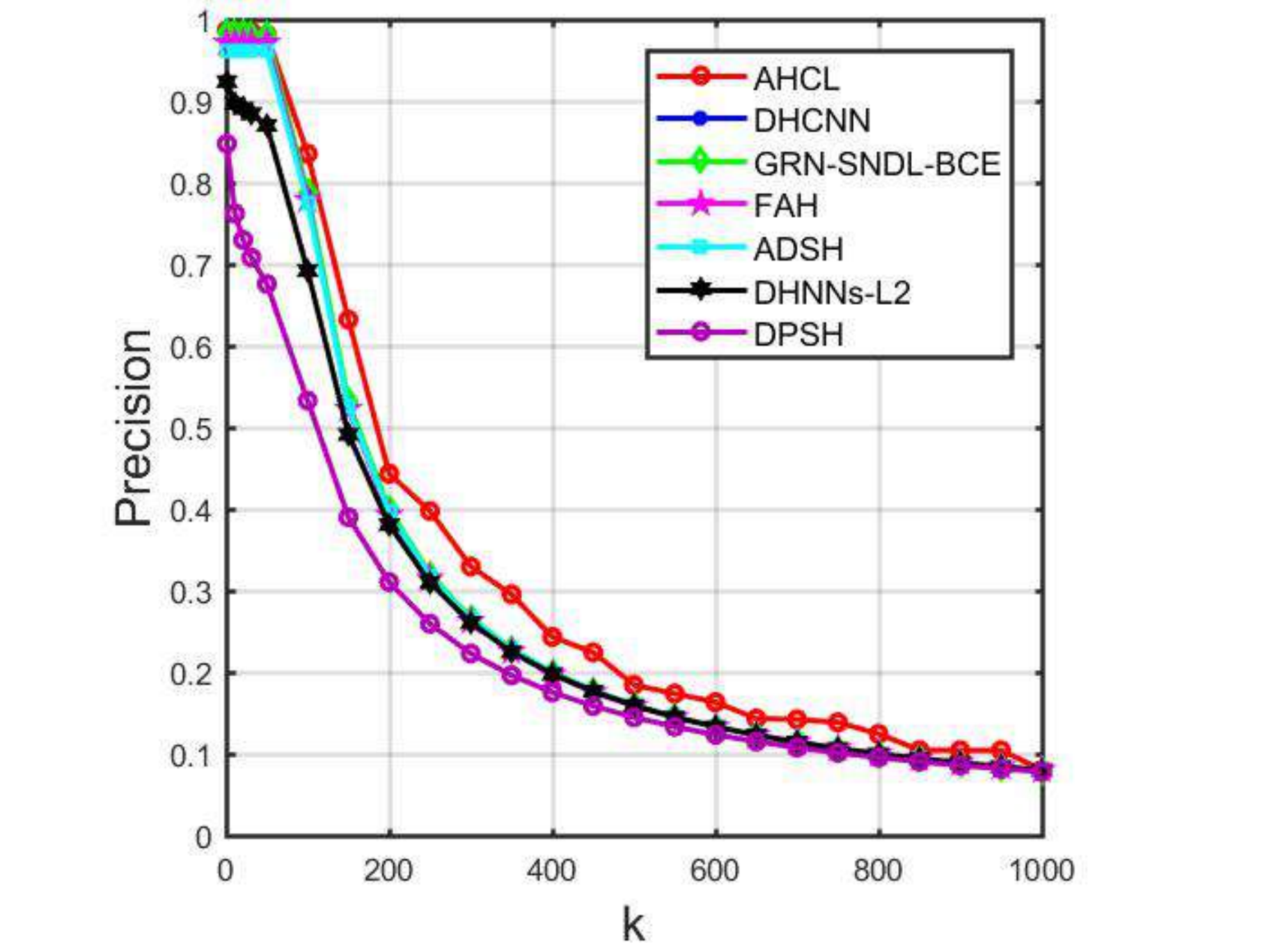}}
\subfigure[]{\includegraphics[width=50mm]{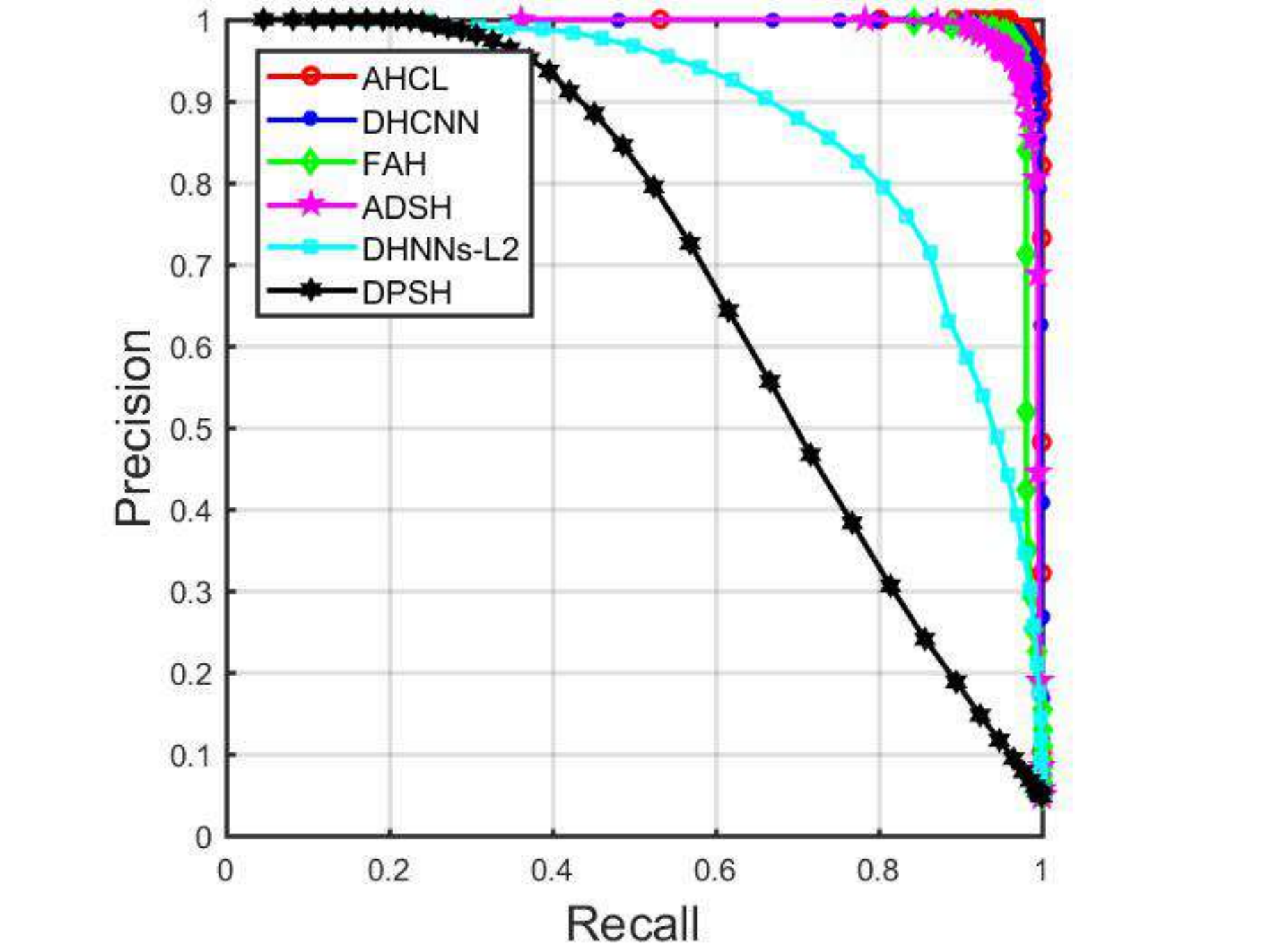}}
\caption{The retrieval results on UCMD with 64-bit hash code. (a) Recall@k; (b) Precision@k; (c) Precision-Recall.}
\label{Curves-UCM21}
\end{center}
\end{figure*}

In the first place, we compare the proposed AHCL with five competitive deep hashing methods, including DPSH, DHNNs-L2, ADSH, FAH, and DHCNN, in terms of qualitative retrieval results. Fig. \ref{fig:retrieval_image} presents query examples with the top-50 retrieved images on the UCMD and AID datasets, where the green rectangle marks the true positives and the red rectangle marks the false positives. Due to the limitation of space, we only visually show the top-12 retrieved images and count the number of true positives out 50 returns. From this figure, we can see that DPSH obtains the worse result results, DHNNs-L2 and FAH perform good for the simple scene (e.g., overpass of UCMD dataset) but perform bad for the complex class (e.g., square of AID dataset). In addition, we also observe that ADSH and DHCNN obtain the competitive retrieval results, but there still exist several false positives in their retrieved images. By contrast, the proposed AHCL returns all true positives for the two examples, which demonstrates the advantages of our method over other compared approaches. In addition, the quantitative retrieval results in terms of MAP are also reported. Table \ref{tab:Results-MAP} shows the MAP of different methods with different hash bits on the three datasets. Considering that GRN-SNDL-BCE is not hashing method, we set the length of feature to 64 and compare it with other hashing approaches with 64 hash bits. From this table, we can obtain the following conclusions: (1) the length of hash codes has a great influence on the retrieval results, and the retrieval performance with short hash bits is generally suboptimal due to their insufficient representation ability; (2) the deep hashing methods are significantly superior to the traditional hashing methods with deep features; (3) by fusing the similarity information and semantic information, the proposed AHCL delivers the better retrieval results than ADSH; (4) GRN-SNDL-BCE obtains the competitive retrieval results, the main reason is the relations between samples are well excavated via a graph relation network and a designed loss function. In addition, GRN-SNDL-BCE adopts the more powerful backbone architecture (i.e., ResNet18 \cite{DRN}) than our pre-trained network (i.e., VGG11 \cite{VGG}); (5) the proposed AHCL obtains the highest MAP values with different hash bits on three datasets.  \par

\begin{figure*}[htbp]
\begin{center}
\subfigure[]{\includegraphics[width=50mm]{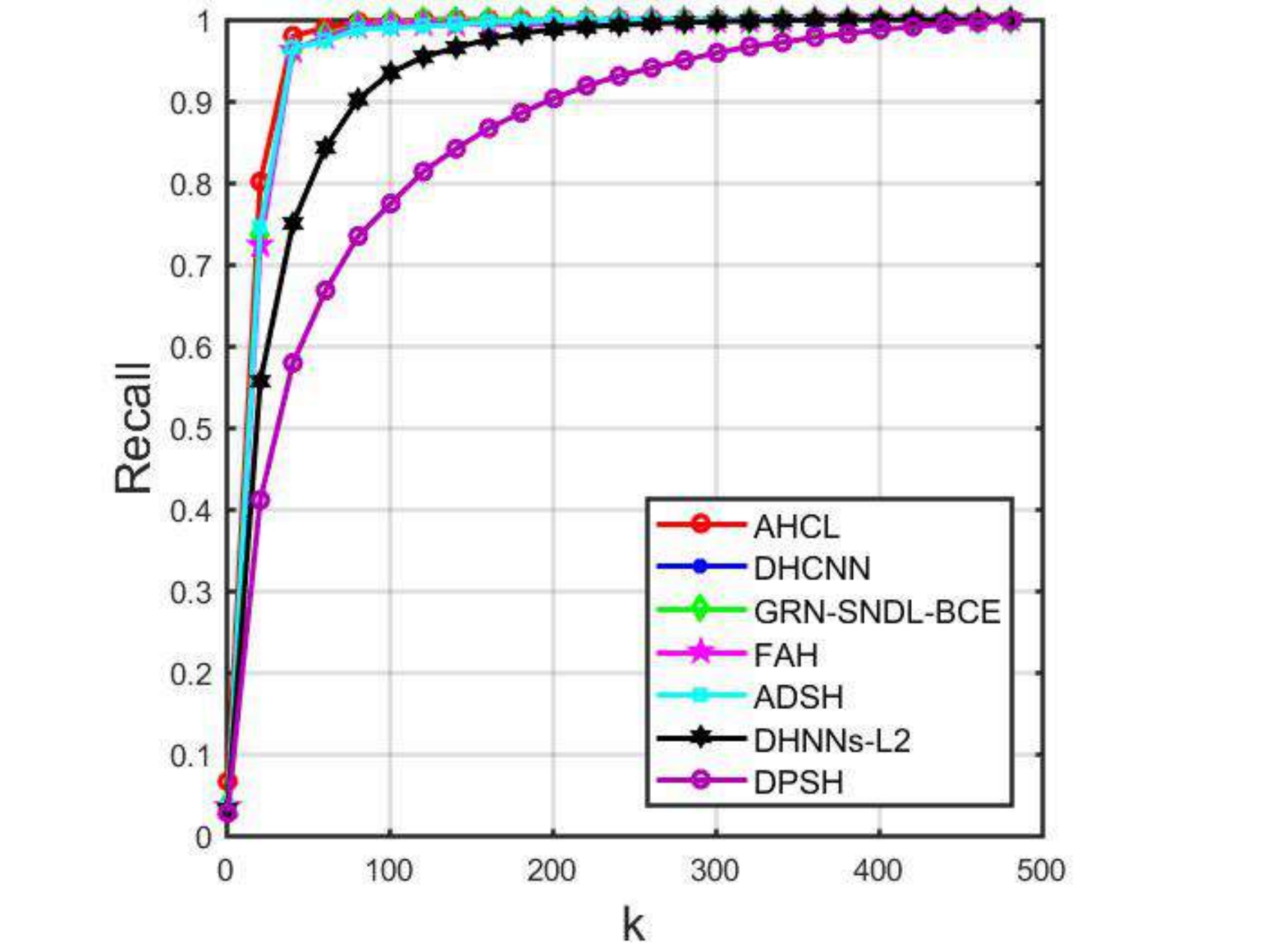}}
\subfigure[]{\includegraphics[width=50mm]{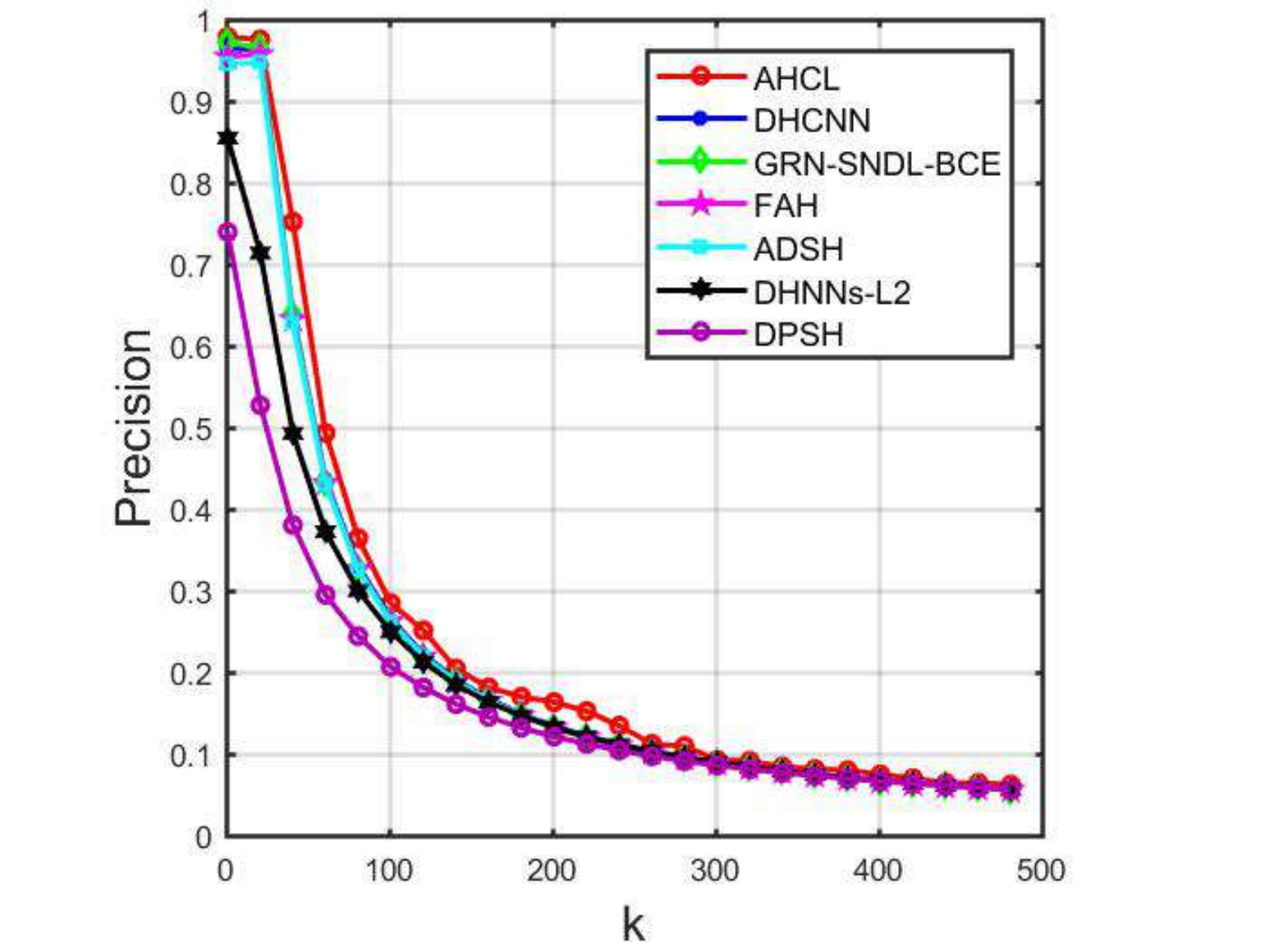}}
\subfigure[]{\includegraphics[width=50mm]{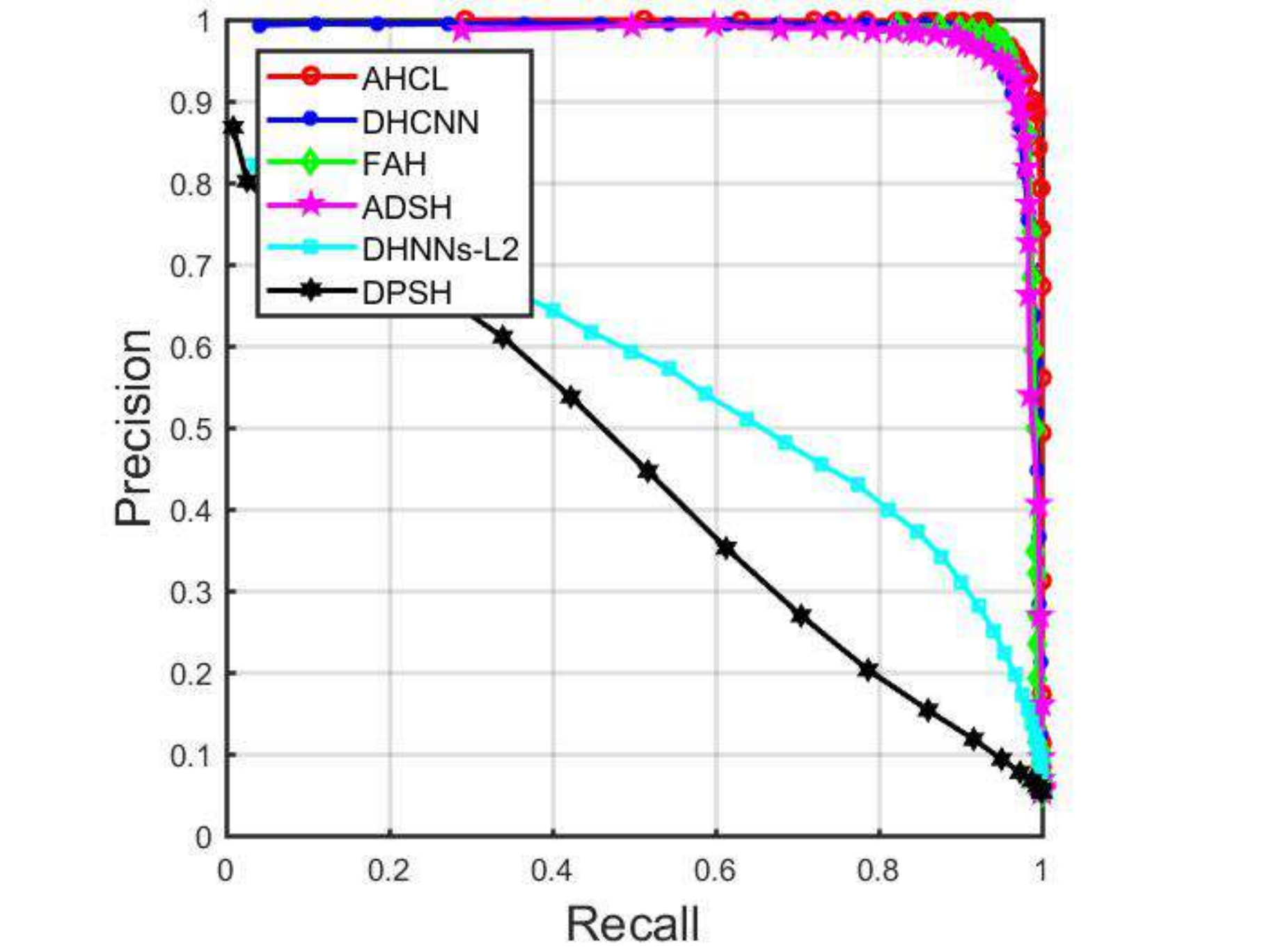}}
\caption{The retrieval results on WHU-RS with 64-bit hash code. (a) Recall@k. (b) Precision@k. (c) Precision-Recall.}
\label{Curves-WHURS19}
\end{center}
\end{figure*}

\begin{figure*}[htbp]
\begin{center}
\subfigure[]{\includegraphics[width=50mm]{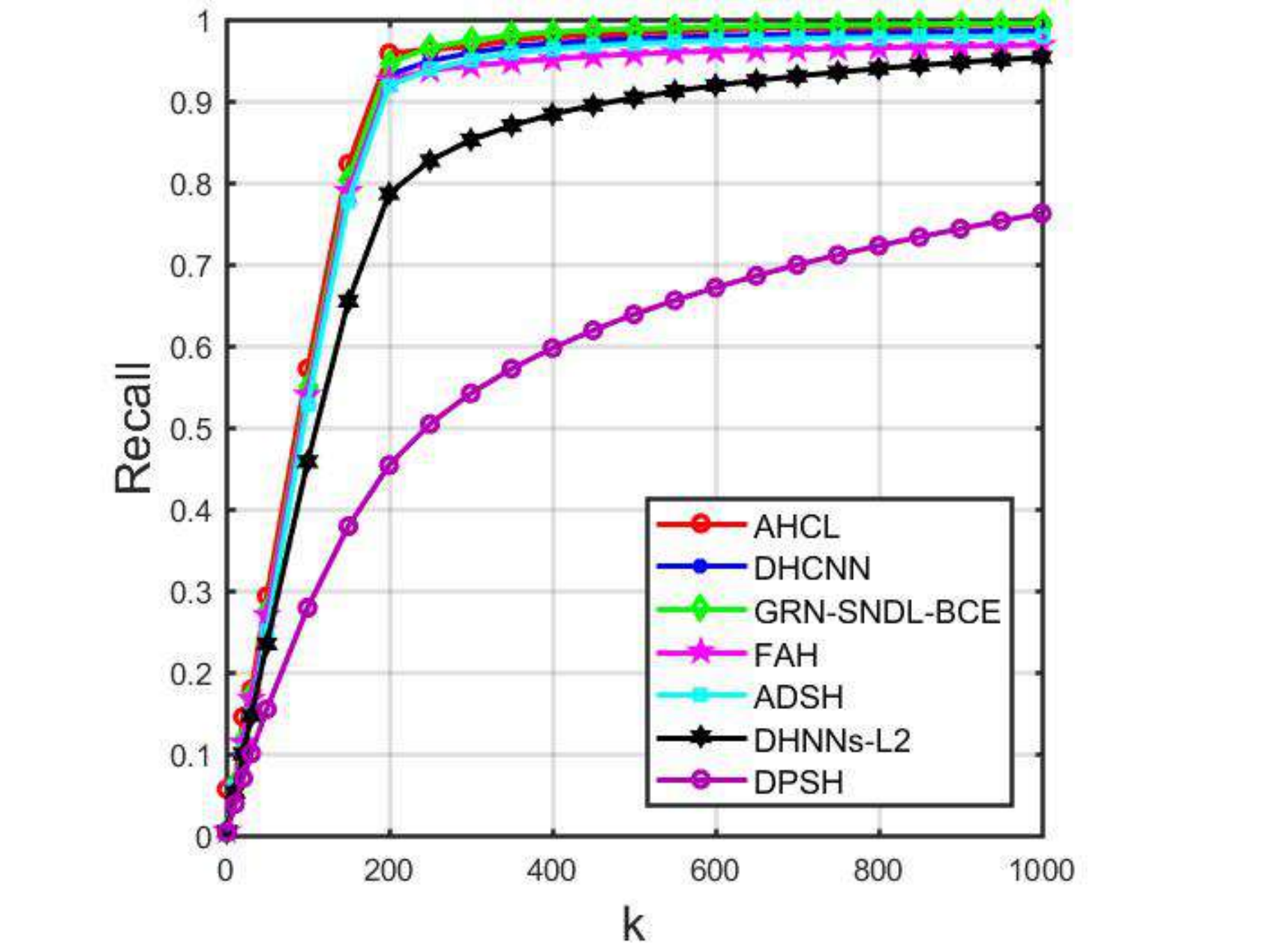}}
\subfigure[]{\includegraphics[width=50mm]{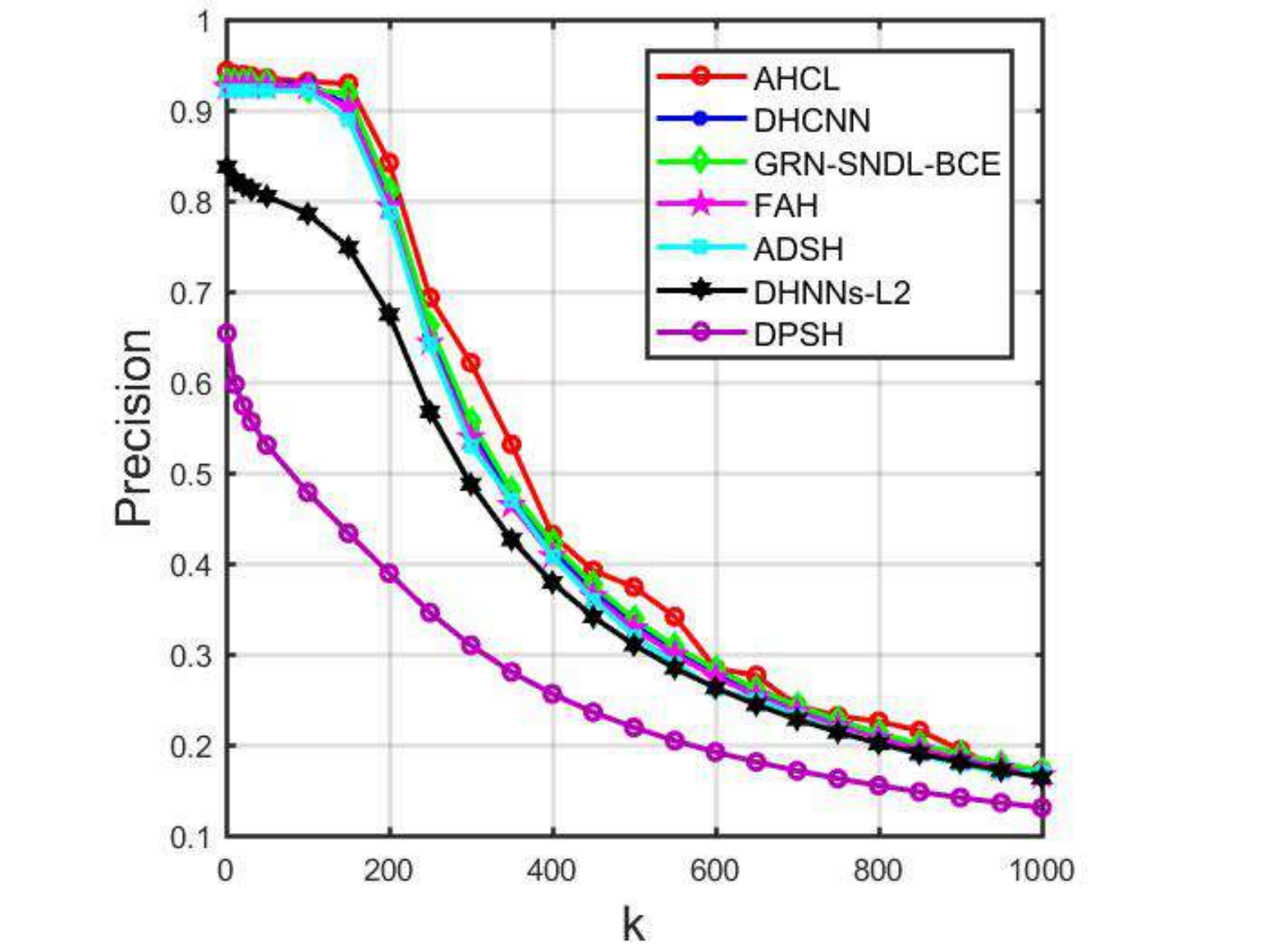}}
\subfigure[]{\includegraphics[width=50mm]{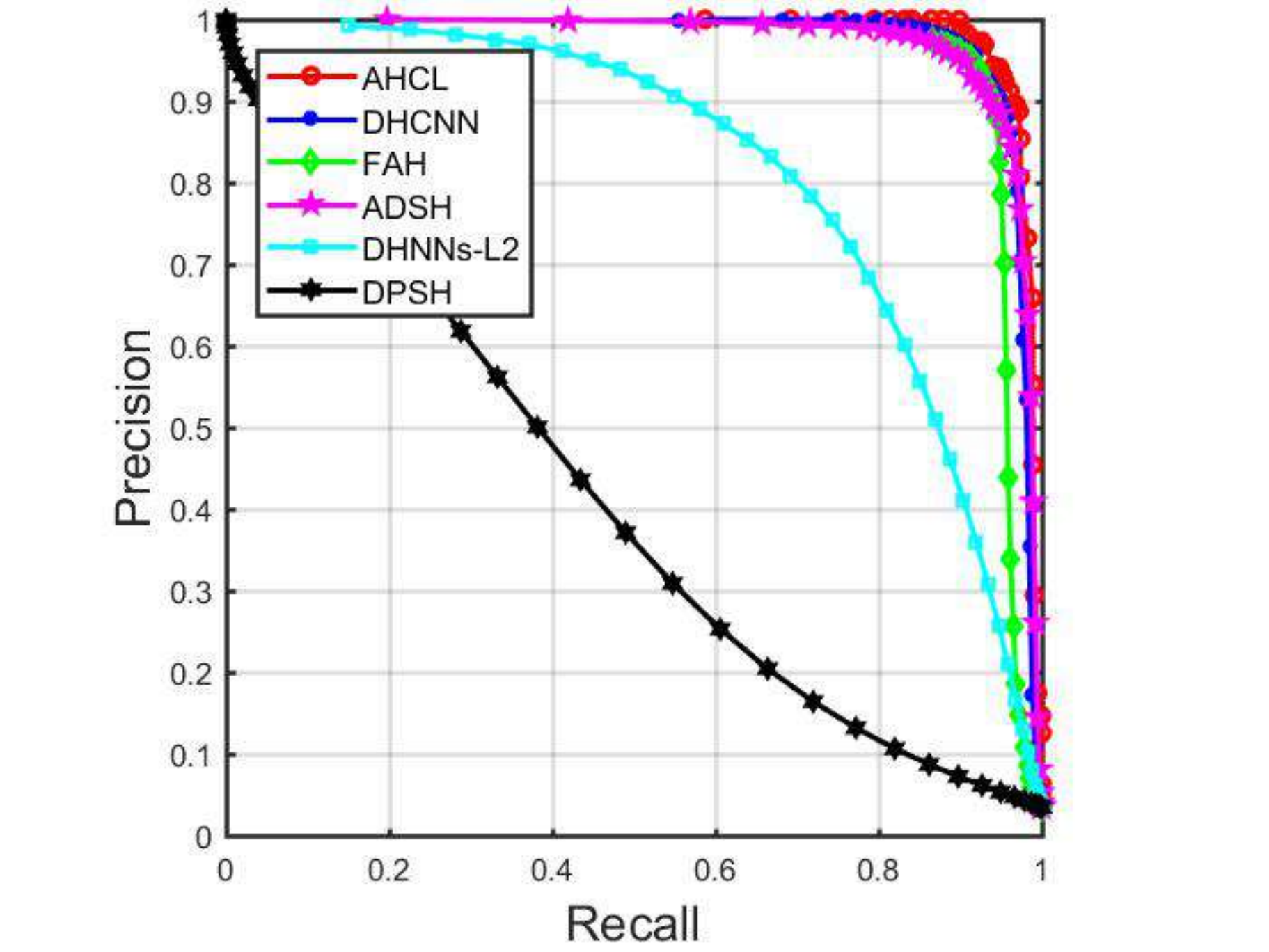}}
\caption{The retrieval results on AID with 64-bit hash code. (a) Recall@k. (b) Precision@k. (c) Precision-Recall.}
\label{Curves-AID30}
\end{center}
\end{figure*}

\begin{table*}[htbp]
\small
\centering
  \caption{Comparison of Running Time (in Seconds) of Different Methods.}
\begin{tabular}{ccccccc}
\hline
\multirow{3}{*}{Methods} & \multicolumn{2}{c}{16 bits}      & \multicolumn{2}{c}{32 bits}  & \multicolumn{2}{c}{64 bits} \\
\cline{2-7}          & Training    & Retrieval         & Training    & Retrieval
                      & Training   & Retrieval    \\
  & time  & time   & time  & time   & time & time   \\
\hline
\textbf{AHCL} (Our method)         & 1243.52  & 24.17     & 1244.45  & 24.09
                       & 1249.75  & 24.26     \\
DHCNN \cite{DHCNN}     & 1064.56  & 67.69     & 1067.45  & 70.43
                       & 1071.78  & 72.85    \\
ADSH \cite{ADSH}       & 1237.29  & 24.21     & 1239.25  & 24.57
                       & 1244.95  & 24.28     \\
DHNNs-L2 \cite{DHNN}   & 1049.56  & 69.49     & 1045.45  & 74.57
                       & 1058.78  & 78.32    \\
DPSH \cite{DPSH}       & 1047.20  & 75.05     & 1049.45  & 75.04
                       & 1056.78  & 78.76      \\
\hline
\end{tabular}
\label{Tab:time-cost-AID}
\end{table*}

\begin{figure*}[htbp]
\begin{center}
\subfigure[UCMD]{\includegraphics[width=45mm]{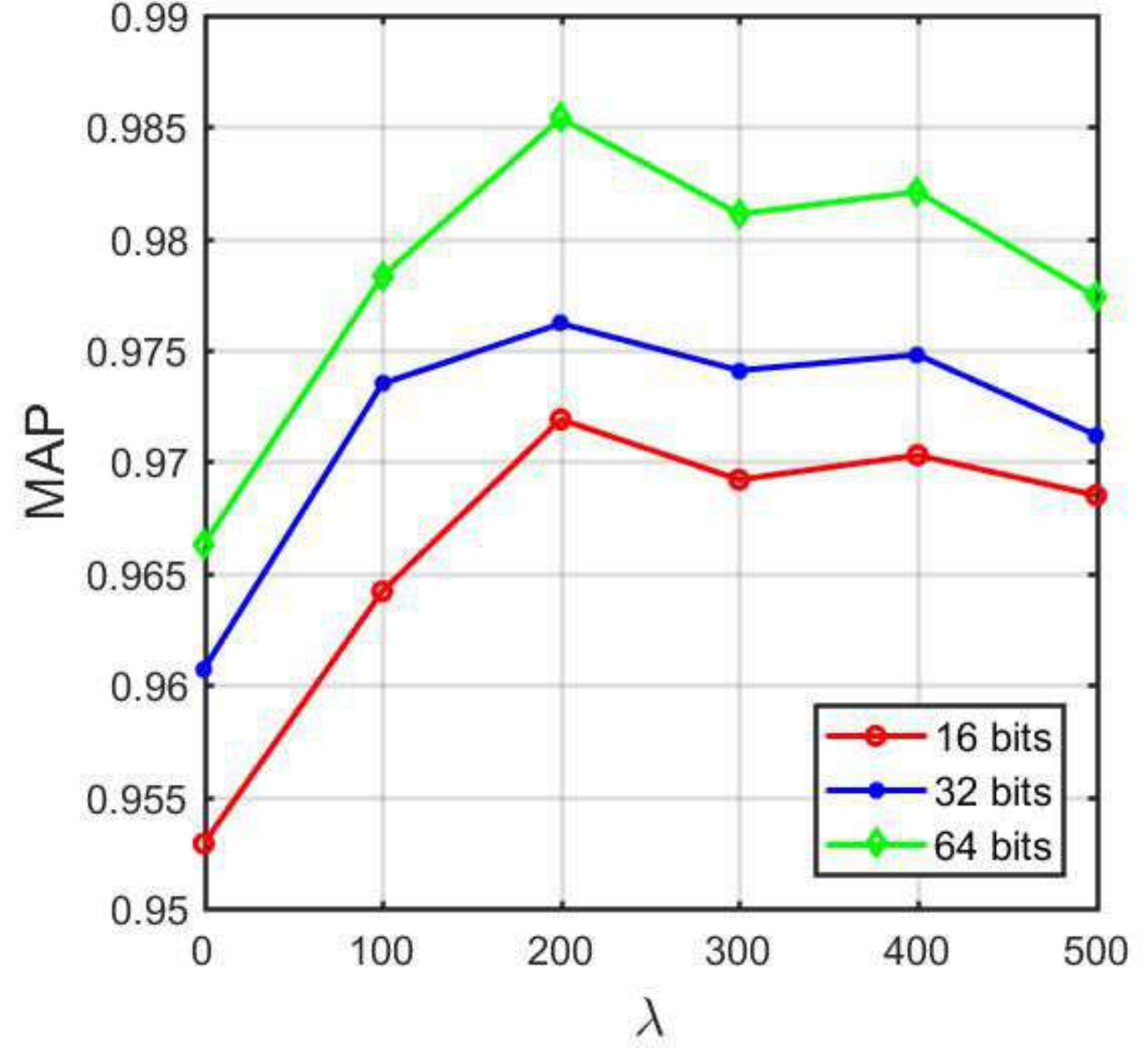}}
\subfigure[WUH-RS] {\includegraphics[width=45mm]{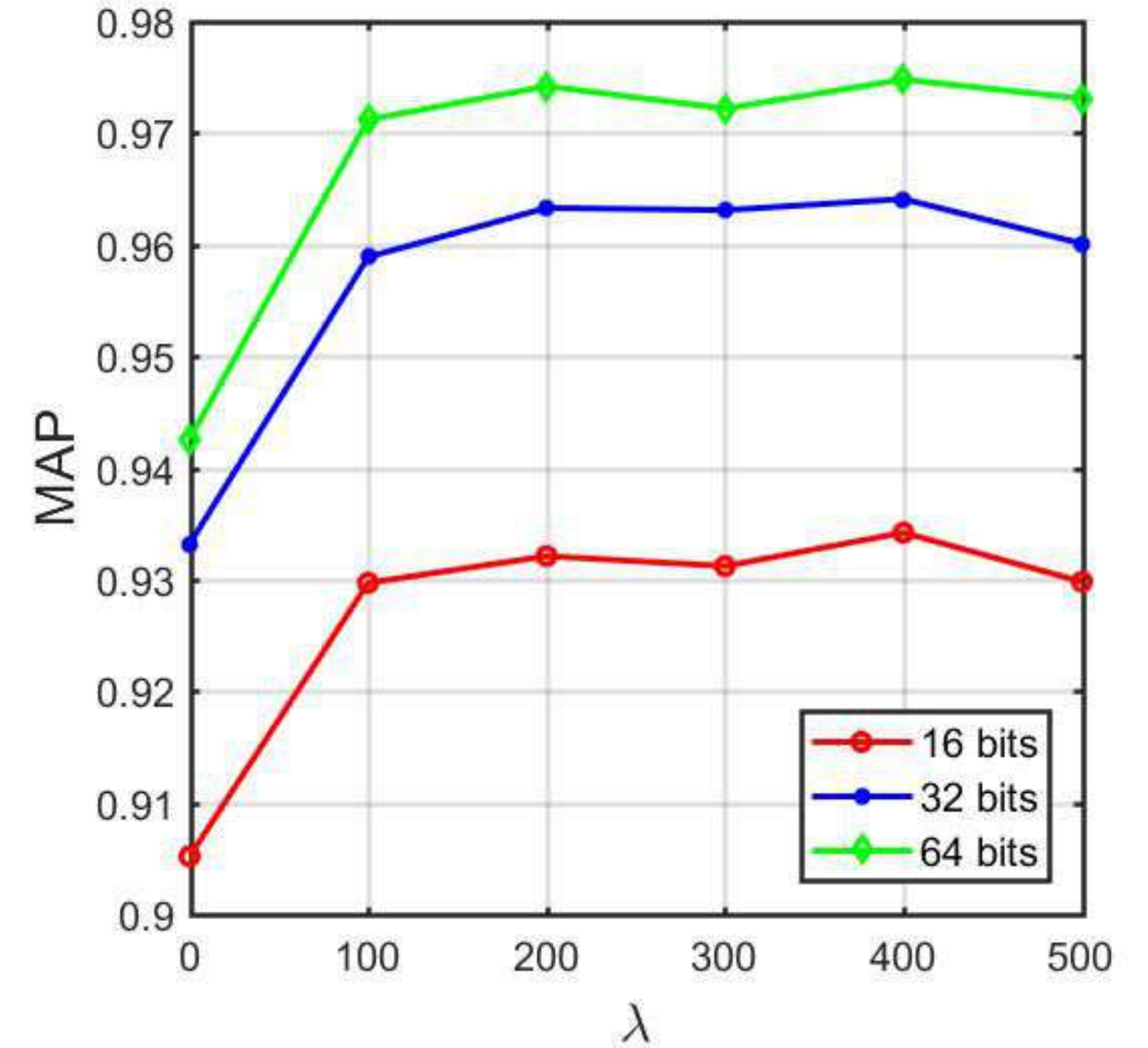}}
\subfigure[AID]{\includegraphics[width=45mm]{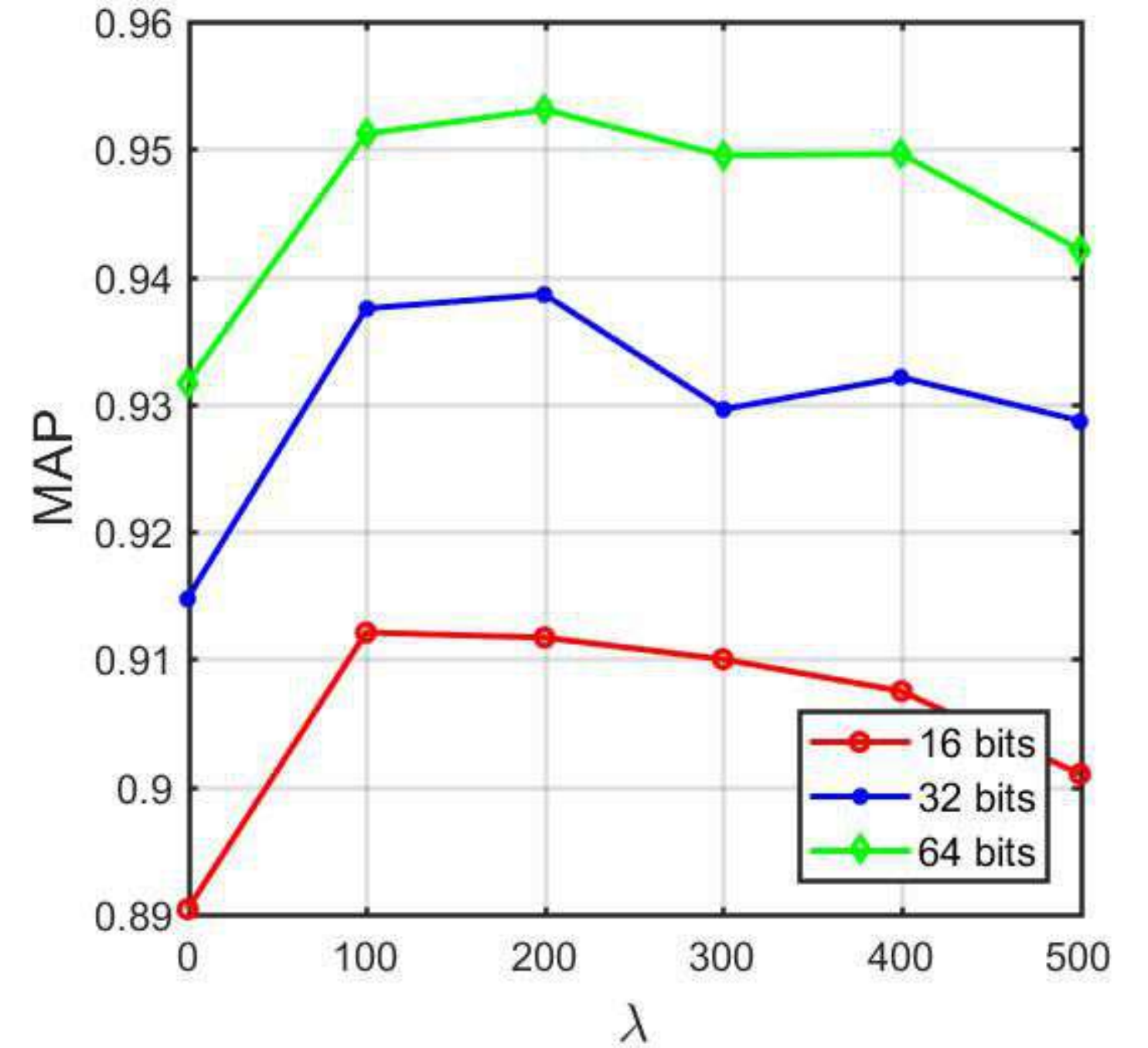}}
\caption{The effects of $\lambda$ on MAP}
\label{lambda-map}
\end{center}
\end{figure*}

\begin{figure*}[htbp]
\begin{center}
\subfigure[UCMD]{\includegraphics[width=49mm]{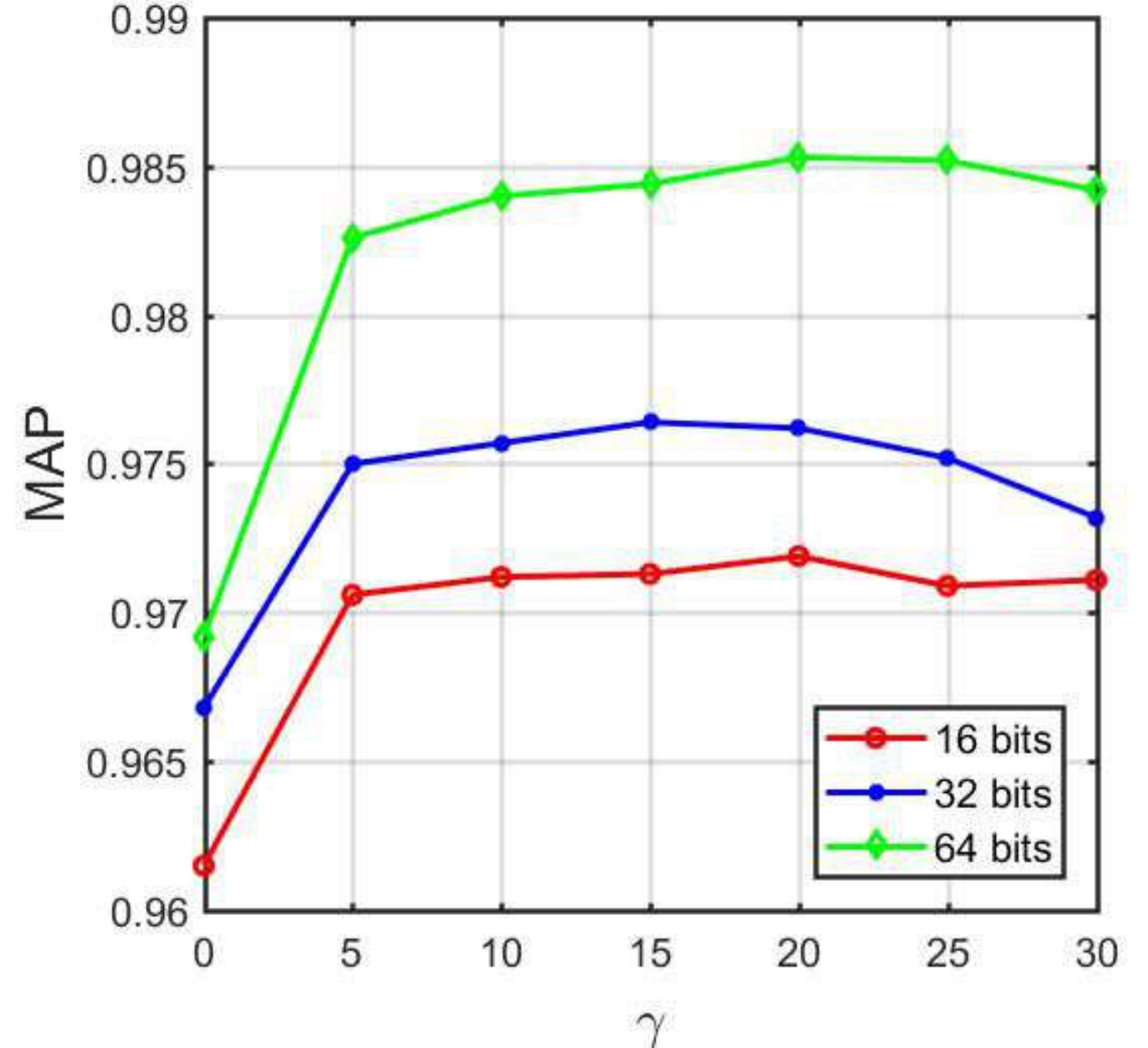}}
\subfigure[WUH-RS] {\includegraphics[width=49mm]{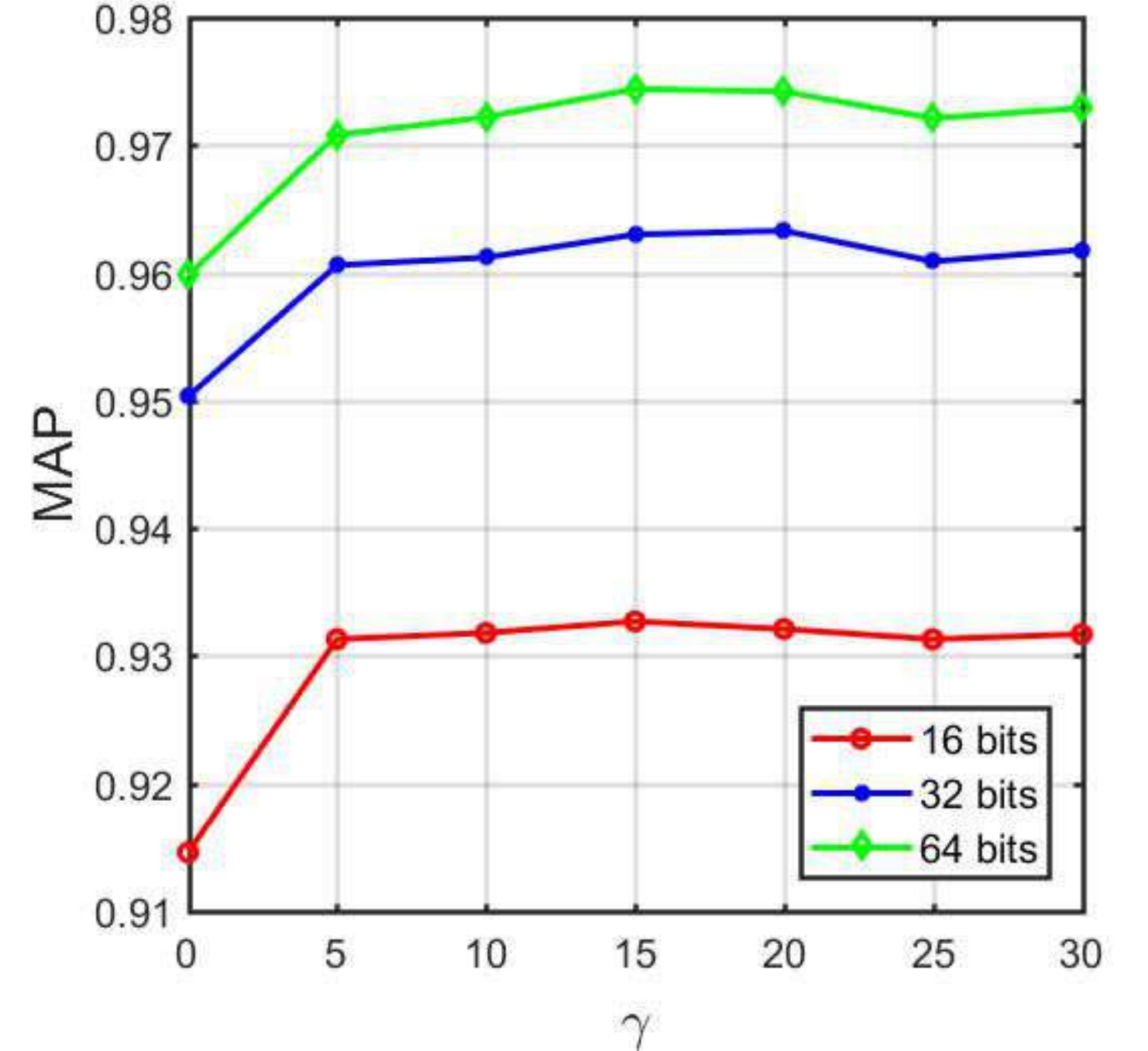}}
\subfigure[AID]{\includegraphics[width=49mm]{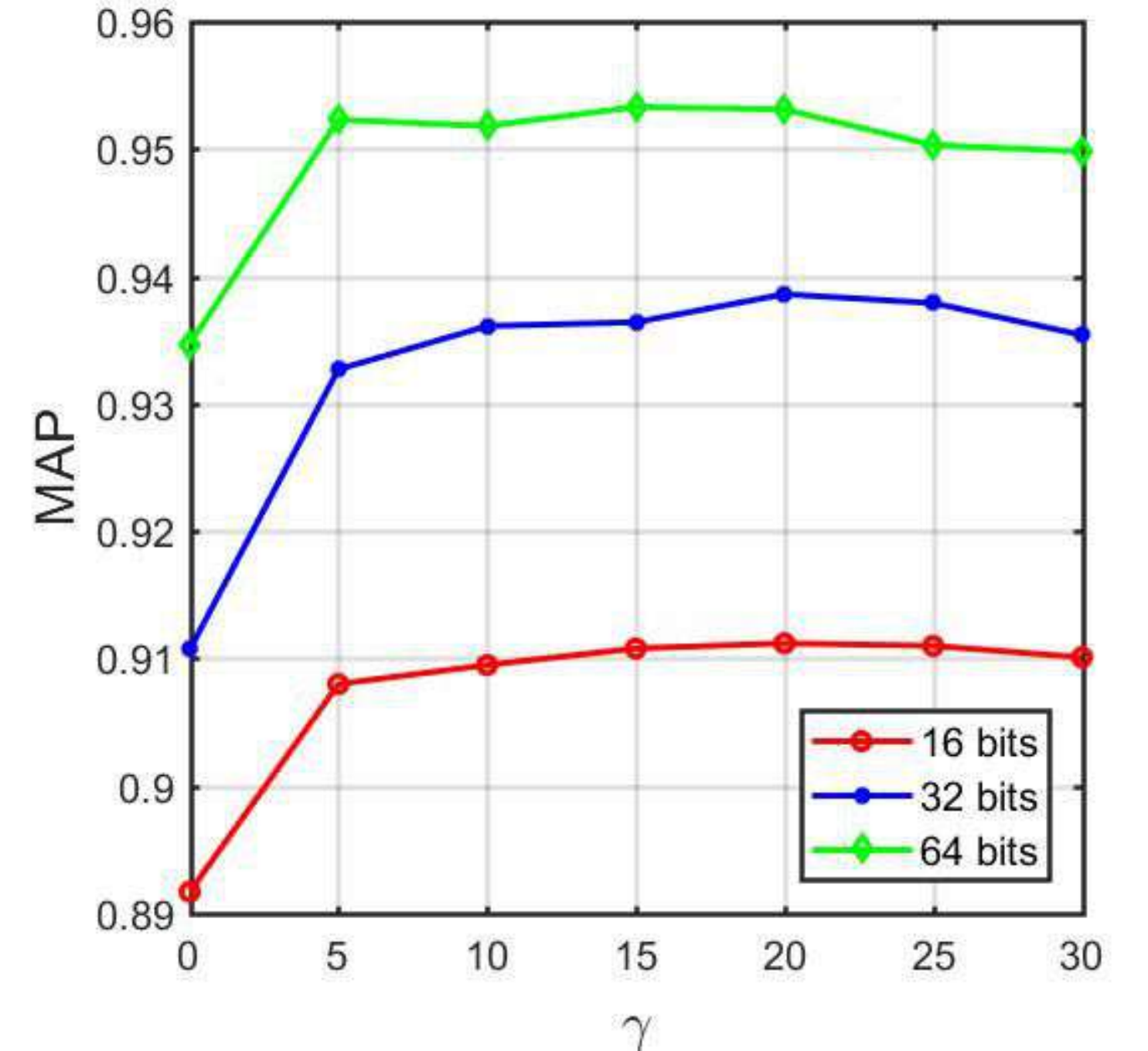}}
\caption{The effects of $\gamma$ on MAP}
\label{gamma-map}
\end{center}
\end{figure*}
Apart from the MAP metric, we also report other three important metrics, i.e., Recall@k, Precision@k, and Precision-Recall curves. In this part, we exclude the traditional hashing methods with deep features (i.e., KSH-CNN \cite{KSH} and SELVE-CNN \cite{SELVE}) due to their poor performance. In addition, considering that the Precision-Recall curves are based on Hamming radius in this paper, thus, the Precision-Recall curve of GRN-SNDL-BCE is not compared with others. Figs.\ref{Curves-UCM21}-\ref{Curves-AID30} show the corresponding retrieval results of different methods on the three datasets, where the hash bit is set to 64. As can be seen from Figs. \ref{Curves-UCM21}-\ref{Curves-AID30}, DPSH and DHNNS-L2 show poor retrieval performance. On the contrary, other methods have achieved satisfactory retrieval results. In addition, the proposed AHCL method has obtained the higher retrieval values than other compared methods in most cases.  \par

\subsection{Effects of Different Training Samples on MAP}

\begin{figure*}
\begin{center}
\subfigure[]{\includegraphics[width=51mm]{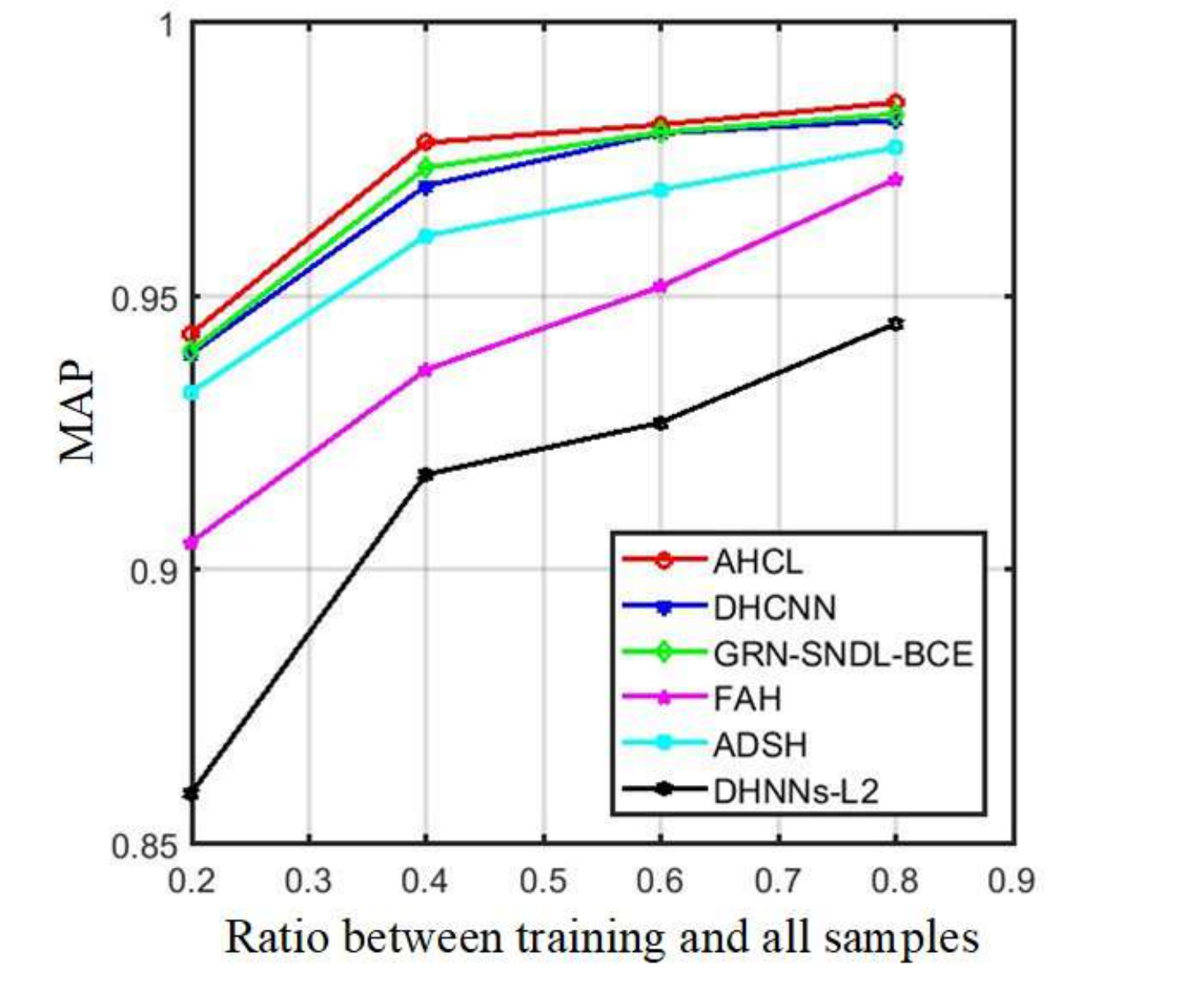}}
\subfigure[]{\includegraphics[width=51mm]{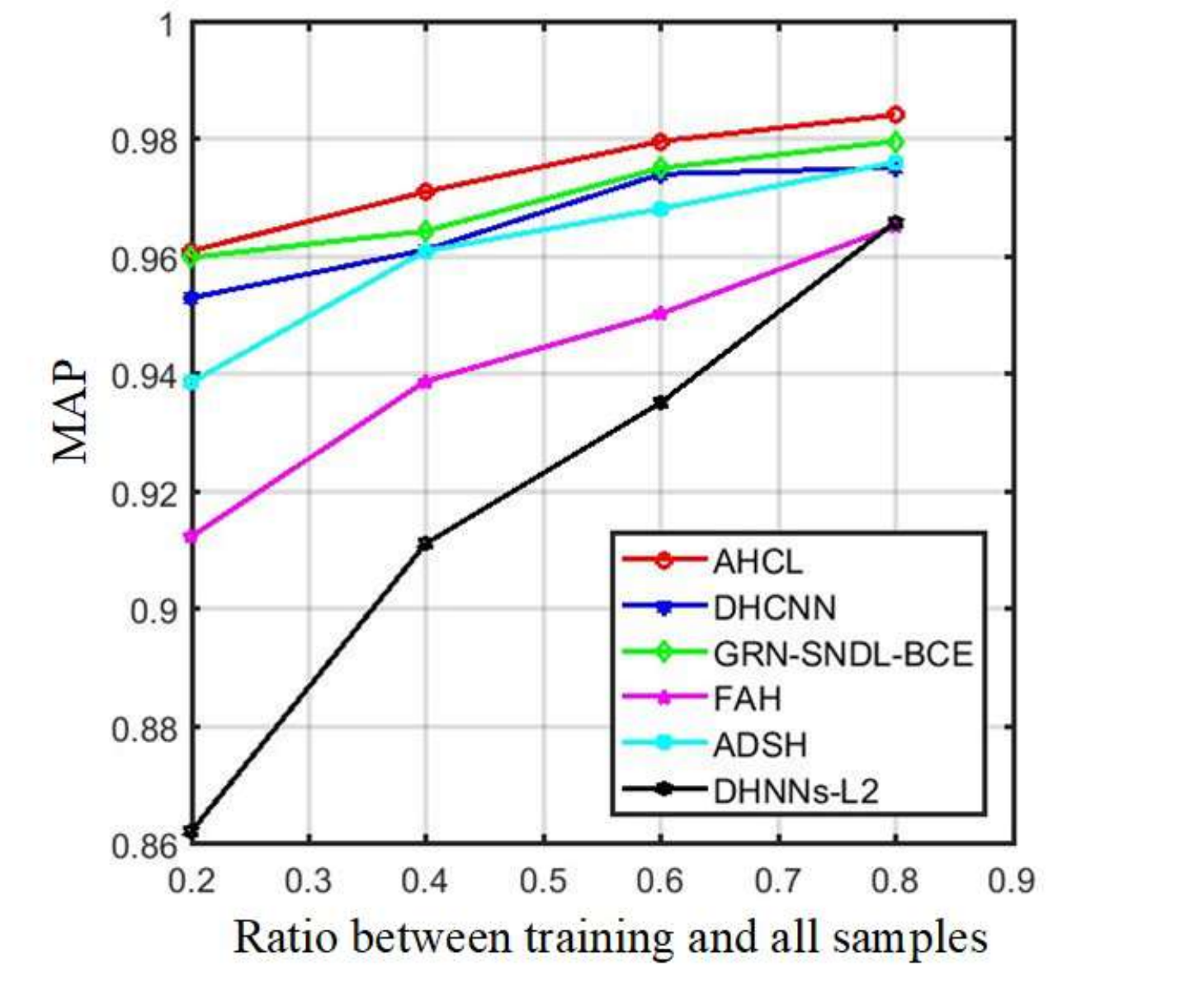}}
\subfigure[]{\includegraphics[width=51mm]{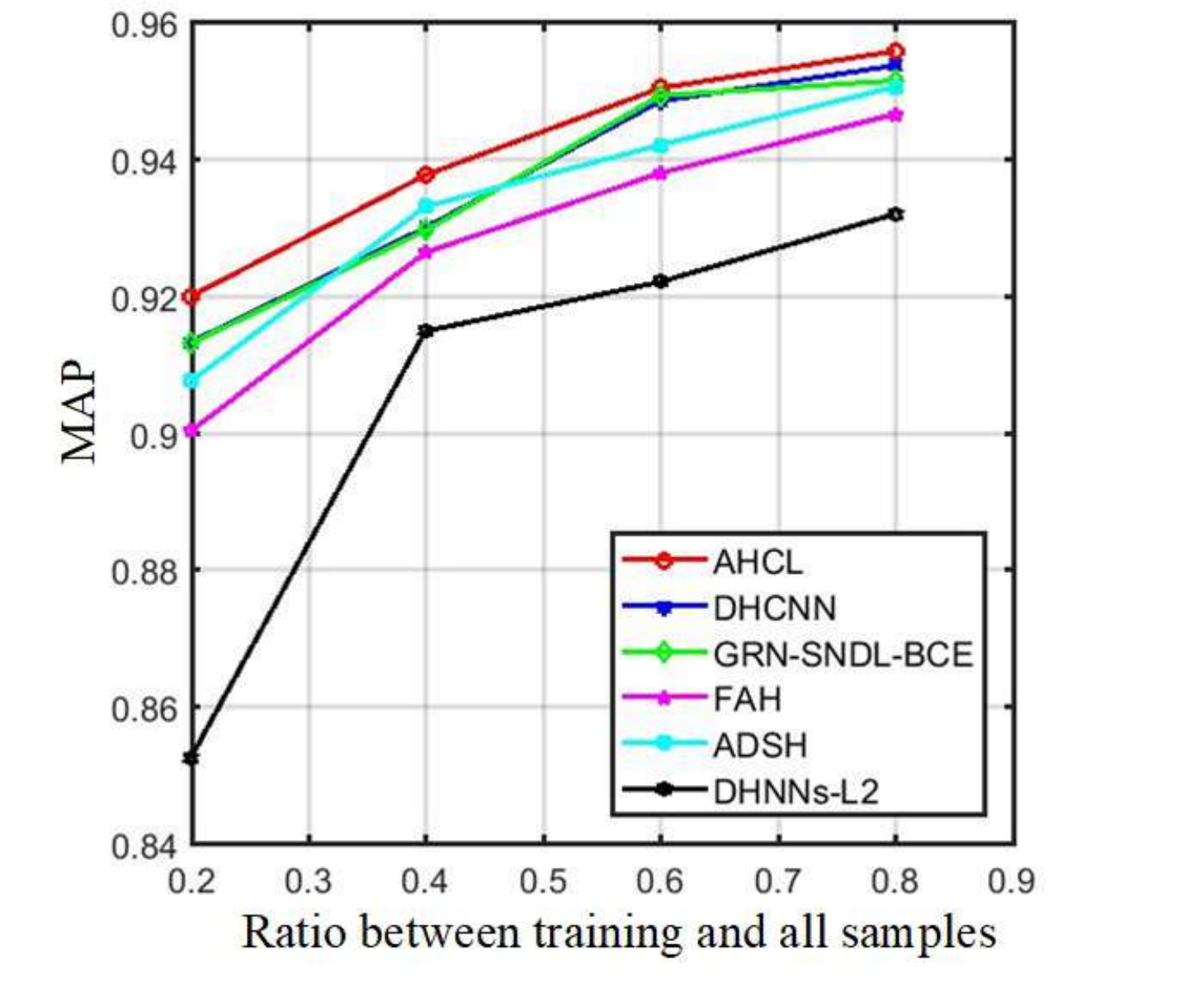}}
\caption{The retrieval results under different number of training samples on the (a) UCMD, (b) WHU-RS, and (c) AID.}
\label{map-samples}
\end{center}
\end{figure*}

In this section, we conduct experiments to analyze the effect of different number of training samples on retrieval result under 64-bit hash codes. Here, we only compare the proposed AHCL with GRN-SNDL-BCE \cite{GRN-SNDL-BCE}, FAH\cite{FAH}, ADSH \cite{ADSH}, and DHNNs-L2 \cite{DHNN}, the rest of compared approaches are excluded due to the poor retrieval results obtained by these methods. The ratio between training and all samples are set to 0.2, 0.4, 0.6, and 0.8 for three datasets.   \par

The retrieval results in terms of MAP are shown in Fig. \ref{map-samples}. From this figure, we can see that DHNNs-L2 is very sensitive to the number of training samples. When small amount of training samples are available, the retrieval results of DHNNs-L2 on three datasets dramatically decrease. By contrast, the MAP values of other methods steadily rise with the number of training samples increasing. Although DHCNN and GRN-SNDL-BCE obtain the competitive results, we can still see that the proposed AHCL delivers the highest MAP values for all separation scenarios on three datasets. Based on the above analyses, we can conclude that the proposed AHCL can achieve satisfactory retrieval performance under different separation scenarios of samples, and at the same time, exhibit advantage over other compared methods to some extent.   \par

\subsection{Computing Time}
In addition to the quantitative metrics, retrieval efficiency is also an important factor when designing a retrieval algorithm. Thus, we further compare the proposed AHCL method with four deep hashing methods in terms of running time. The experiments are performed on the AID dataset and the training ratio is set to 0.5 per class. Table \ref{Tab:time-cost-AID} shows the comparison of the training and retrieval time of different methods with 16, 32, and 64 hash bits.
From this table, we can see that the training time of DHCNN is slightly higher than that of DHNNS-L2 and DPSH. At the same time, the proposed AHCL also takes more training time than that of ADSH. The main reason for the above experimental phenomenon is that AHCL and DHCNN add a semantic layer after the hash layer to consider the semantic information of images. Therefore, these two methods need more time to train the additional parameter layer. In addition, we can also observe that the retrieval time of the AHCL and ADSH methods is much lower than that of other three symmetric hashing methods, which verifies the efficiency of the asymmetric strategy for hash code learning.   \par

\subsection{Parameter Analysis}
\label{sec:analysis}
As can be seen from Equation (\ref{J}), the proposed AHCL method contains two important hyper-parameters, i.e., $\lambda$ and $\gamma$. The variable $\lambda$ is used to constrain the representation error between two hash code representations for query images. The variable $\gamma$ is used to balance the similarity loss and semantic loss. In the following parts, the effects of the above two hyper-parameters on MAP values are analyzed in detail.    \par

\subsubsection{The effects of $\lambda$ on MAP}

In order to analyze the effect of $\lambda$ on the retrieval performance, the $\gamma$ value is set to 20 according to experience \cite{DHCNN}. Figure \ref{lambda-map} shows the changing curve of the MAP values with the increase of $\lambda$ on the UCMD, WHU-RS, and AID datasets, where the hash bits $K$ are set to 16, 32 and 64, respectively. As can be seen from this figure, when the hash bit $K$ increases, the MAP values improve significantly. In addition, when $\lambda=0$, MAP values reach the lowest value, the main reason is that the objective function cannot effectively constrain the approximation error between the hash-like codes and the hash codes to be enough small. For both UCMD and AID datasets, the best MAP values are achieved when $\lambda=200$ for all hash bits. Although for the WHU-RS dataset, the MAP values under $\lambda=200$ do not reach the maximum values, the results at this time are still very close to the maximum values. Thus, we set $\lambda=200$ as the optimal value for the three datasets.  \par

\subsubsection{The effects of $\gamma$ on MAP}

Before analyzing the effects of $\gamma$ on MAP, the $\lambda$ value is set to 200 for all three datasets. Fig. \ref{gamma-map} shows the effects of $\gamma$ on the MAP values. From this figure, we can see that when $\gamma$ approaches to 0, the MAP values significantly decrease for all scenarios. The main reason for the above experimental phenomenon is that objective function under the condition of $\gamma=0$ only considers the similar information between images, while ignoring the semantic information of each image. When $\gamma>0$, the MAP values increase significantly and become stable with the increase of $\gamma$. Through the above observation, it is found that the proposed method can obtain satisfactory retrieval results on all datasets when $\gamma$ equals 20.  \par

\section{Conclusions}
\label{sec:conclusions}
Currently, deep hashing-based RSIR methods attempt to learn one hash function for both query and database samples in a symmetric way. Specifically, the hash codes of query and database remote sensing images are all obtained by binarizing the output of the network. However, it is typically time-consuming to generate the hash codes of huge database images. To this end, we proposed a novel asymmetric deep hashing method for fast RSIR. In more detail, the hash codes of query images are obtained via the feedback computation of the deep hashing network, while the hash codes of database images are directly learned by solving the objective function. The proposed asymmetric strategy improves the generation efficiency of hash codes, which is vital for the large-scale retrieval task. In addition, the designed loss function simultaneously exploits the semantic information and similarity information of images to enhance the ability of feature representation. Finally, the experimental results validate the superiority of the proposed method over the compared approaches. \par

\ifCLASSOPTIONcaptionsoff
  \newpage
\fi




%





\bibliographystyle{IEEEbib}
\bibliography{manuscript}

\begin{IEEEbiography}[{\includegraphics[width=1in,height=1.25in,clip,keepaspectratio]
{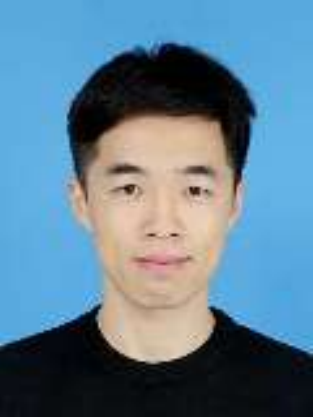}}]{Weiwei Song}
received the B.S. degree in automation from Southwest Minzu University, Chengdu, China, in 2015, and the Ph.D. degree in control science and engineering from Hunan University, Changsha, China, in 2021. From November 2018 to November 2019, he was a visiting Ph.D. student under the supervision of Prof. J\'on Atli Benediktsson with the Department of Electrical and Computer Engineering, University of Iceland, Reykjavik, Iceland, supported by the China Scholarship Council.  \par
He is a currently Post-Doctoral Researcher with the Department of Mathematics and Theories, Peng Cheng Laboratory, Shenzhen. His research interests include deep learning, machine learning, and remote sensing and their applications.    \par
\end{IEEEbiography}

\begin{IEEEbiography}[{\includegraphics[width=1in,height=1.25in,clip,keepaspectratio]
{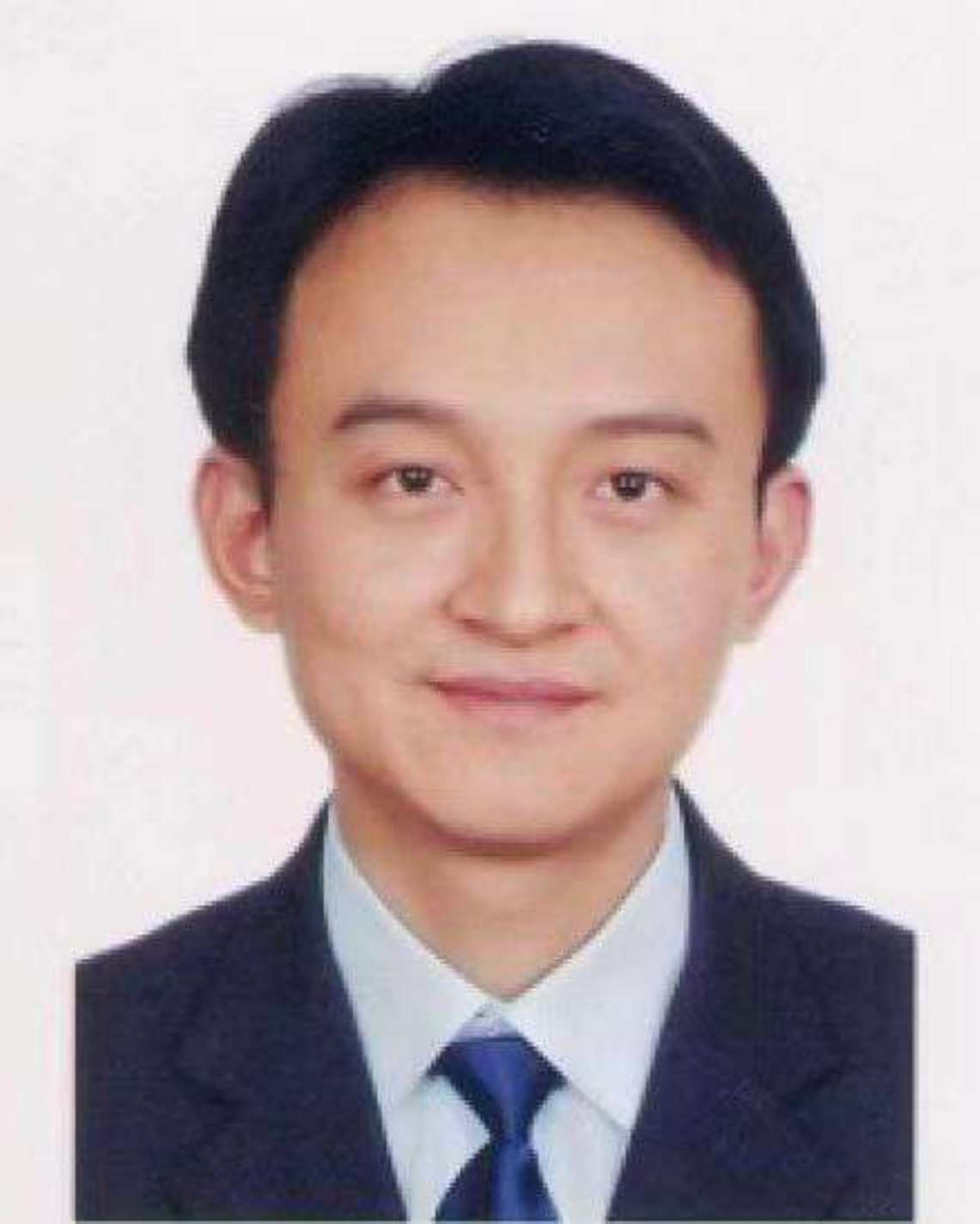}}]{Zhi Gao}
received the B.Eng. and the Ph.D. degrees from Wuhan University, China, in 2002 and 2007,
respectively. In 2008, he joined the Interactive and Digital Media Institute, National University of Singapore (NUS), as a Research Fellow (A) and the Project Manager. In 2014, he joined the Temasek Laboratories in NUS (TL@NUS) as a Research Scientist (A) and the Principal Investigator. He is currently working as a Full Professor with the School of Remote Sensing and Information Engineering, Wuhan University. He has published more than 70 research papers on top journals and conferences, such as IJCV, IEEE TRANSACTIONS ON PATTERN ANALYSIS AND MACHINE INTELLIGENCE (TPAMI), IEEE TRANSACTIONS ON INDUSTRIAL ELECTRONICS (TIE), IEEE TRANSACTIONS ON GEOSCIENCE AND REMOTE SENSING (TGRS), IEEE TRANSACTIONS ON INTELLIGENT TRANSPORTATION SYSTEMS (TITS), ISPRS Journal of Photogrammetry and Remote Sensing (JPRS), Neurocomputing, IEEE TRANSACTIONS ON CIRCUITS AND SYSTEMS FOR VIDEO TECHNOLOGY (TCSVT), CVPR, ECCV, ACCV, and BMVC. His research interests include computer vision, machine learning, and remote sensing and their applications. In particular, he has strong interests in vision for intelligent systems and intelligent system-based vision. Since 2019, he has been supported by the Distinguished Professor Program of Hubei Province and the National Young Talent Program, China. He serves as an Associate Editor for the journal Unmanned Systems.
\end{IEEEbiography}

\begin{IEEEbiography}[{\includegraphics[width=1in,height=1.25in,clip,keepaspectratio]
{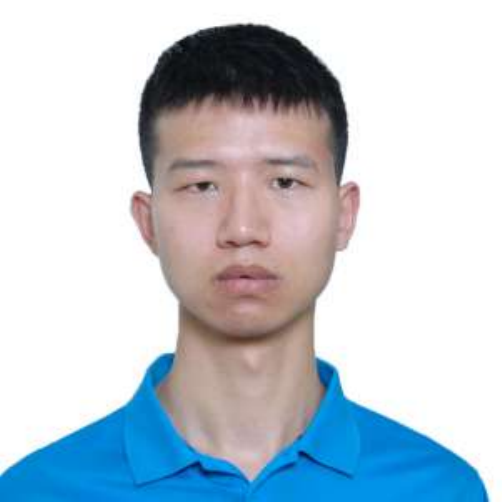}}]{Renwei Dian}
(S'16-M'20) received the B.S. degree from
Wuhan University of Science and Technology, Wuhan, China, in 2015, and the Ph.D. degree from Hunan University, Changsha, China, in 2020, respectively. He is currently a postdoctor in the College of Electrical and Information Engineering, Hunan University.

From November 2017 to November 2018, he is a visiting Ph.D. student with  University of Lisbon, Lisbon, Portugal, supported by the China Scholarship Council. He was a finalist for the Best Student Paper Award at the  International Geoscience and Remote Sensing Symposium (IGARSS) 2018. He was awarded the Fellowship of China National Postdoctoral Program for Innovative Talents in 2020 and  Excellent Doctoral Dissertation by China Society of Image and Graphics in 2020.
His research interests include hyperspectral image super-resolution, image fusion, tensor decomposition, and deep learning.   More information can be found in
his homepage {https://sites.google.com/view/renweidian/.}
\end{IEEEbiography}

\begin{IEEEbiography}[{\includegraphics[width=1in,height=1.25in,clip,keepaspectratio]
{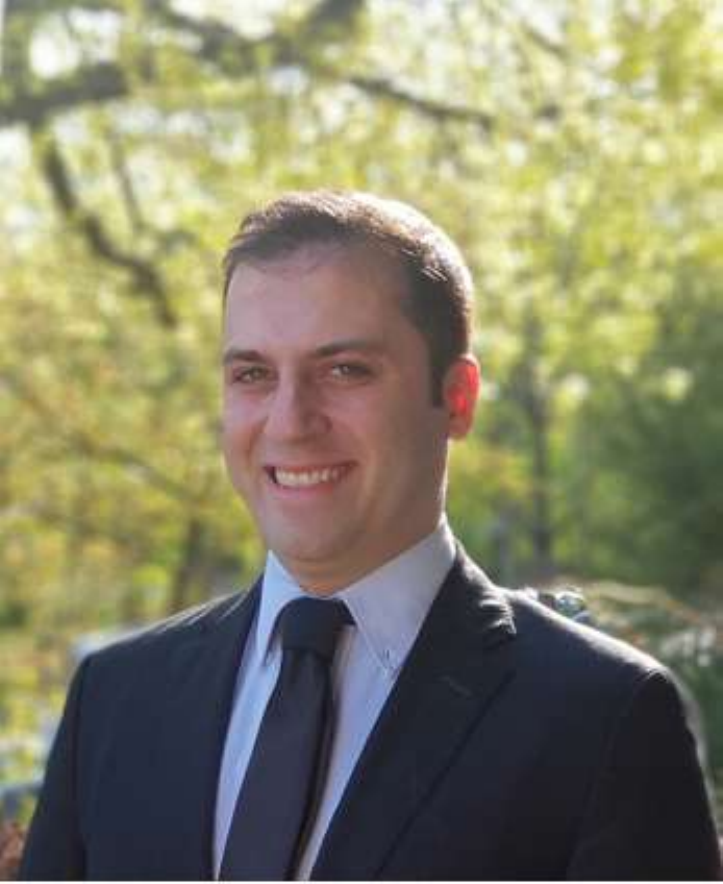}}]{Pedram Ghamisi}
(S¡¯12, M¡¯15, SM¡¯18) graduated with a Ph.D. in electrical and computer engineering at the University of Iceland in 2015. He works as (1) the head of the machine learning group at Helmholtz-Zentrum Dresden-Rossendorf (HZDR), Germany and (2) research professor and group leader of AI4RS at the Institute of Advanced Research in Artificial Intelligence (IARAI), Austria. He is a co-founder of VasoGnosis Inc. with two branches in San Jose and Milwaukee, the USA.  \par
He was the co-chair of IEEE Image Analysis and Data Fusion Committee (IEEE IADF) between 2019 and 2021. Dr. Ghamisi was a recipient of the IEEE Mikio Takagi Prize for winning the Student Paper Competition at IEEE International Geoscience and Remote Sensing Symposium (IGARSS) in 2013, the first prize of the data fusion contest organized by the IEEE IADF in 2017, the Best Reviewer Prize of IEEE Geoscience and Remote Sensing Letters in 2017, and the IEEE Geoscience and Remote Sensing Society 2020 Highest-Impact Paper Award. His research interests include interdisciplinary research on machine (deep) learning, image and signal processing, and multisensor data fusion. He is an associate editor of IEEE JSTARS and IEEE GRSL. For detailed info, please see {http://pedram-ghamisi.com/}. \par
\end{IEEEbiography}

\begin{IEEEbiography}[{\includegraphics[width=1in,height=1.25in,clip,keepaspectratio]
{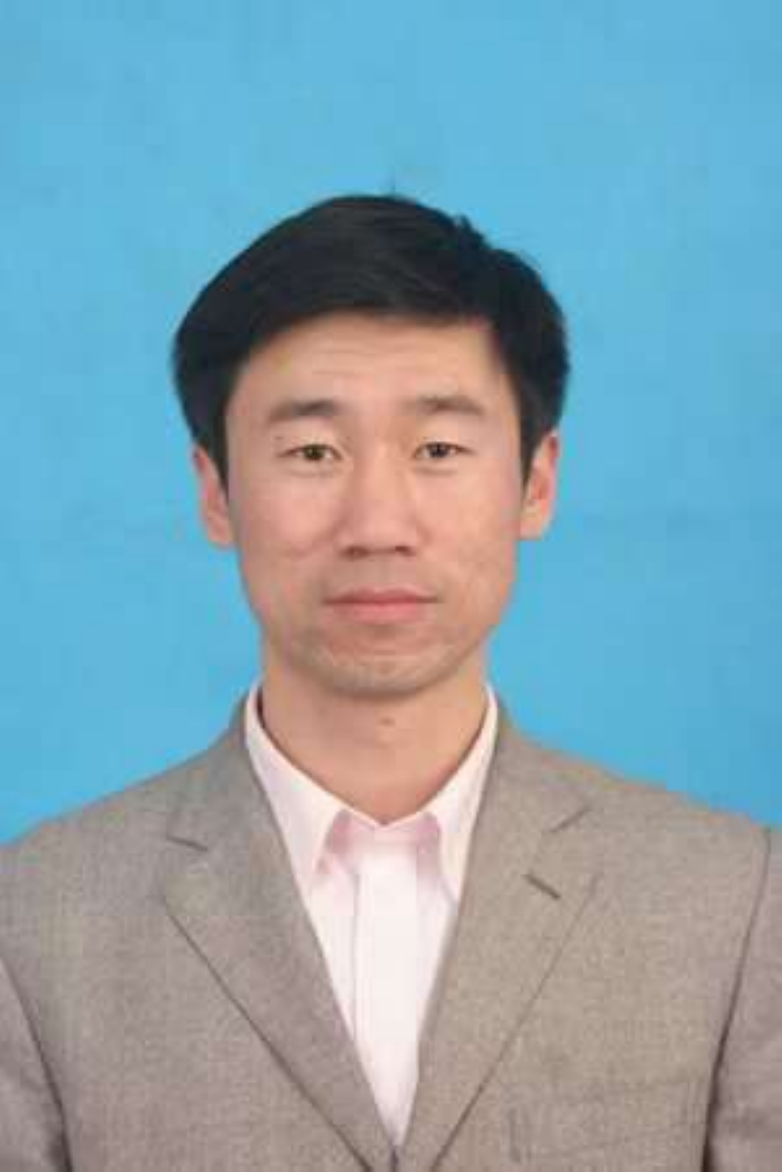}}]{Yongjun Zhang}
received the B.S. degree in Geodesy, the M.S. degree in Geodesy and Surveying Engineering, and the Ph.D. degree in Geodesy and Photography from Wuhan University, Wuhan, China, in 1997, 2000, and 2002, respectively. He is currently Dean of the School of Remote Sensing and Information Engineering, Wuhan University, Wuhan, China. Since 2006, he has been a Full Professor of the School of Remote Sensing and Information Engineering, Wuhan University. From 2014 to 2015, he was a Senior Visiting Fellow with the Department of Geomatics Engineering at University of Calgary, Canada. From 2015 to 2018, he was a Senior Scientist at Environmental Systems Research Institute, Inc. (Esri), USA. He has published more than 150 research articles and one book. He holds 25 Chinese patents and 26 copyright registered computer software. His research interests include aerospace and low-attitude photogrammetry, image matching, combined block adjustment with multisource data sets, object information extraction and modelling with artificial intelligence, integration of LiDAR point clouds and images, and 3-D city model reconstruction.
\end{IEEEbiography}

\begin{IEEEbiography}[{\includegraphics[width=1in,height=1.25in,clip,keepaspectratio]
{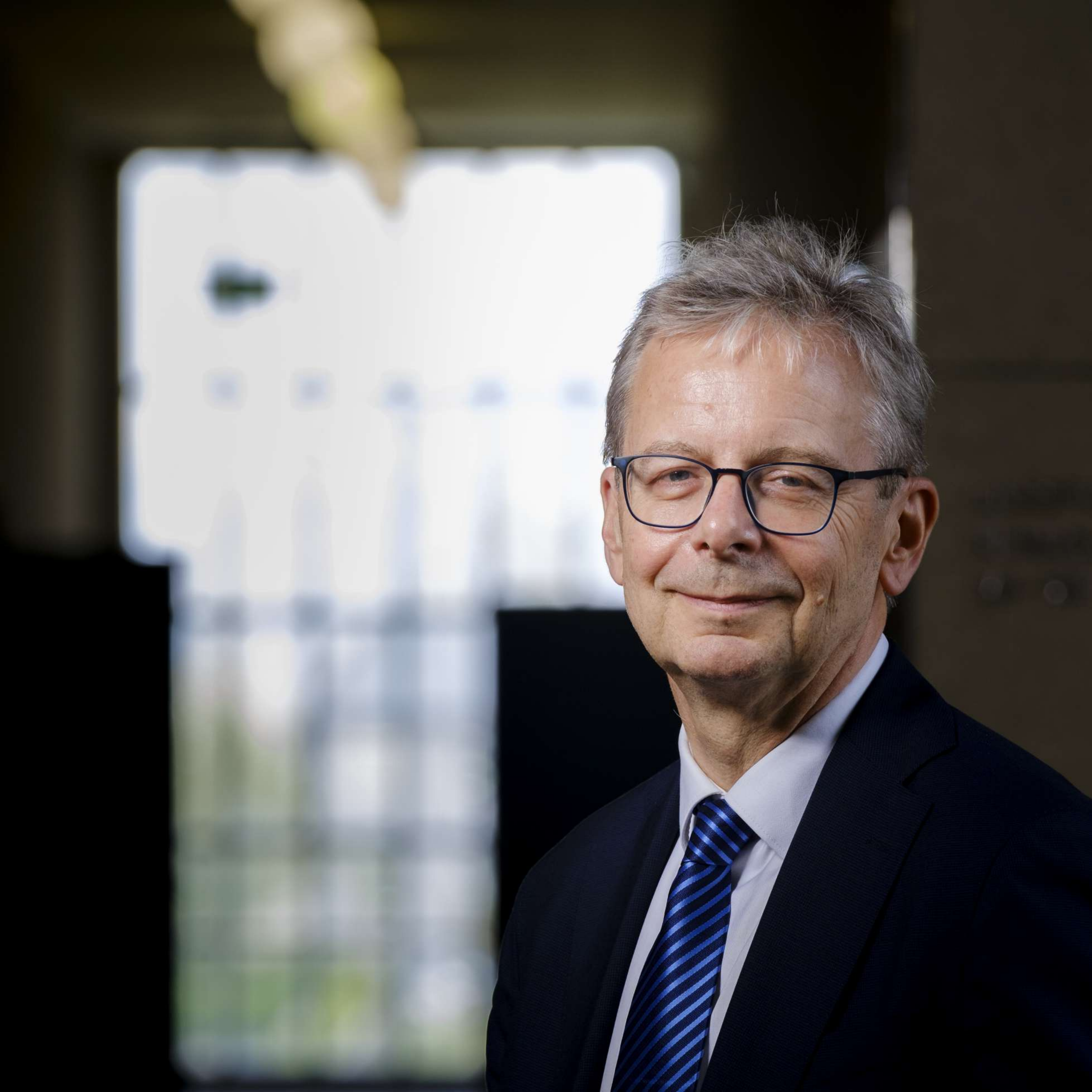}}]{J\'on Atli Benediktsson}
received the Cand.Sci. degree in electrical engineering from the University of Iceland, Reykjavik, in 1984, and the M.S.E.E. and Ph.D. degrees in electrical engineering from Purdue University, West Lafayette, IN, in 1987 and 1990, respectively. Since July 1, 2015 he is the President and Rector of the University of Iceland. From 2009 to 2015 he was the Pro Rector of Science and Academic Affairs and Professor of Electrical and Computer Engineering at the University of Iceland. His research interests are in remote sensing, biomedical analysis of signals, pattern recognition, image processing, and signal processing, and he has published extensively in those fields. Prof. Benediktsson is a Highly Cited Researcher (Clarivate Analysis, 2018-2021). He was the 2011-2012 President of the IEEE Geoscience and and Remote Sensing Society (GRSS) and has been on the GRSS AdCom since 2000. He was Editor in Chief of the IEEE Transactions on Geoscience and Remote Sensing (TGRS) from 2003 to 2008 and has served as Associate Editor of TGRS since 1999, the IEEE Geoscience and Remote Sensing Letters since 2003 and IEEE Access since 2013. He is currently Senior Editor of the Proceedings of the IEEE, is on the International Editorial Board of the International Journal of Image and Data Fusion, the Editorial Board of Remote Sensing, and was the Chairman of the Steering Committee of IEEE Journal of Selected Topics in Applied Earth Observations and Remote Sensing (J-STARS) 2007-2010. Prof. Benediktsson is a co-founder of the biomedical start up company Oxymap (www.oxymap.com). He is a Fellow of the IEEE and a Fellow of SPIE. Prof. Benediktsson was a member of the 2014 IEEE Fellow Committee. He received the Stevan J. Kristof Award from Purdue University in 1991 as outstanding graduate student in remote sensing. In 1997, Dr. Benediktsson was the recipient of the Icelandic Research Council's Outstanding Young Researcher Award, in 2000 he was granted the IEEE Third Millennium Medal, in 2004, he was a co-recipient of the University of Iceland's Technology Innovation Award, in 2006 he received the yearly research award from the Engineering Research Institute of the University of Iceland, in 2007 he received the Outstanding Service Award from the IEEE GRSS, in 2020 the IEEE GRSS Education Award, in 2018 the IEEE GRSS David Landgrebe Award and in 2016 the OECE Award from the School of ECE, Purdue University. He was co-recipient of the 2012 IEEE Transactions on Geoscience and Remote Sensing Paper Award and in 2013 he was co-recipient of the IEEE GRSS Highest Impact Paper Award. In 2013 he received the IEEE/VFI Electrical Engineer of the Year Award. In 2016 and 2018, he was a co-recipient of the International Journal of Image and Data Fusion Best Paper Award. In 2021, he was honored as a recipient of the Order of the Falcon from the President of Iceland. He is a member of Academia Europea, the Association of Chartered Engineers in Iceland (VFI), Societas Scinetiarum Islandica and Tau Beta Pi.
\end{IEEEbiography}
\end{document}